\documentclass[10pt,twocolumn,letterpaper]{article}

\usepackage{iccv}
\usepackage{times}
\usepackage{epsfig}
\usepackage{graphicx}
\usepackage{amsmath}
\usepackage{amssymb}
\usepackage{subcaption}
\usepackage{multirow}

\usepackage[pagebackref=true,breaklinks=true,letterpaper=true,colorlinks,bookmarks=false]{hyperref}
\usepackage{cleveref}

\iccvfinalcopy 

\def\iccvPaperID{5437} 

\begin{document}

\title{Temporal Attentive Alignment for Large-Scale Video Domain Adaptation}

\author{Min-Hung Chen$^1$\thanks{Work partially done as a SIE intern} \hspace{1em}
Zsolt Kira$^1$ \quad
Ghassan AlRegib$^1$ \quad
Jaekwon Yoo$^2$ \quad
Ruxin Chen$^2$ \quad
Jian Zheng$^3$\footnotemark[1]\\
$^1$Georgia Institute of Technology \quad
$^2$Sony Interactive Entertainment LLC \quad
$^3$Binghamton University\\
}

\maketitle

\begin{abstract}
   Although various image-based domain adaptation (DA) techniques have been proposed in recent years, domain shift in videos is still not well-explored. Most previous works only evaluate performance on small-scale datasets which are saturated. Therefore, we first propose two large-scale video DA datasets with much larger domain discrepancy: \textbf{UCF-HMDB$\boldsymbol{_{full}}$} and \textbf{Kinetics-Gameplay}. Second, we investigate different DA integration methods for videos, and show that simultaneously aligning and learning temporal dynamics achieves effective alignment even without sophisticated DA methods. Finally, we propose \textbf{Temporal Attentive Adversarial Adaptation Network (TA$\boldsymbol{^3}$N)}, which explicitly attends to the temporal dynamics using domain discrepancy for more effective domain alignment, achieving state-of-the-art performance on four video DA datasets (e.g. 7.9\% accuracy gain over ``Source only" from 73.9\% to 81.8\% on ``HMDB $\rightarrow$ UCF", and 10.3\% gain on ``Kinetics $\rightarrow$ Gameplay"). The code and data are released at \url{http://github.com/cmhungsteve/TA3N}.
\end{abstract}

\section{Introduction}

\begin{figure}[!t]
\centering
\includegraphics[width=0.475\textwidth]{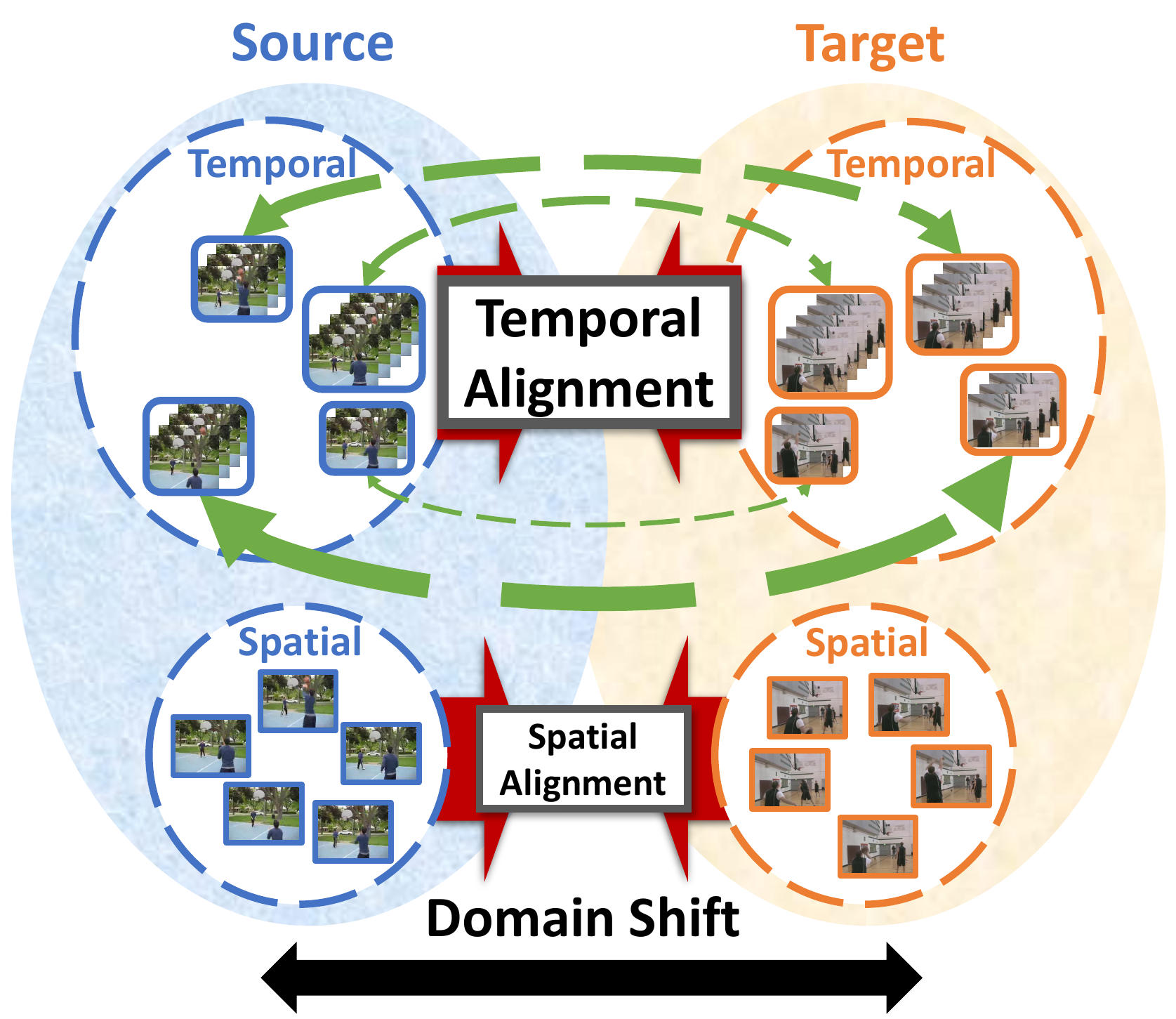}
\caption{An overview of proposed TA$^3$N for video DA. In addition to spatial discrepancy between frame images, videos also suffer from temporal discrepancy between sets of time-ordered frames that contain multiple local temporal dynamics with different contributions to the overall domain shift, as indicated by the thickness of green dashed arrows. 
Therefore, we propose to focus on aligning the temporal dynamics which have higher domain discrepancy using a learned attention mechanism to effectively align the temporal-embedded feature space for videos.
Here we use the action \textit{basketball} as the example.}
\label{fig:overview_video_DA}
\end{figure}

\textit{Domain adaptation (DA)}~\cite{pan2010survey} has been studied extensively in recent years~\cite{csurka2017comprehensive} to address the \textit{domain shift} problem~\cite{quionero2009dataset, peng2018syn2real}, which means the models trained on source labeled dataset do not generalize well to target datasets and tasks.
DA is categorized in terms of the availability of annotations in the target domain. 
In this paper, we focus on the harder unsupervised DA problem, which requires training models that can generalize to target samples without access to any target labels. While many unsupervised DA approaches are able to diminish the distribution gap between source and target domains while learning discriminative deep features~\cite{long2015learning, long2017deep, ganin2015unsupervised, ganin2016domain, li2017revisiting, li2018adaptive, saito2018maximum}, most methods have been developed only for images and not videos.

Furthermore, unlike image-based DA work, there do not exist well-organized datasets to evaluate and benchmark the performance of DA algorithms for videos. 
The most common datasets are \textit{UCF-Olympic} and \textit{UCF-HMDB$_{small}$}~\cite{sultani2014human, xu2016dual, jamal2018deep}, which have only a few overlapping categories between source and target domains. 
This introduces limited domain discrepancy so that a deep CNN architecture can achieve nearly perfect performance even without any DA method (details in \Cref{sec:experimental_results} and \Cref{table:sota_ucf-olympic_ucf-hmdb-small}).
Therefore, we propose two larger-scale datasets to investigate video DA: 1) \textit{UCF-HMDB$_{full}$}: We collect 12 overlapping categories between UCF101~\cite{soomro2012ucf101} and HMDB51~\cite{kuehne2011hmdb}, which is around three times larger than both UCF-Olympic and UCF-HMDB$_{small}$, and contains larger domain discrepancy (details in \Cref{sec:experimental_results} and \Cref{table:sota_ucf-hmdb_full,table:sota_hmdb-ucf_full}). 2) \textit{Kinetics-Gameplay}: We collect from several currently popular video games with 30 overlapping categories with Kinetics-600~\cite{kay2017kinetics, carreira2018kinetics}. This dataset is much more challenging than UCF-HMDB$_{full}$ due to the significant domain shift between the distributions of virtual and real data. 

Videos can suffer from domain discrepancy along both the spatial and temporal directions, bringing the need of alignment for embedded feature spaces along both directions, as shown in \Cref{fig:overview_video_DA}. 
However, most DA approaches have not explicitly addressed the domain shift problem in the temporal direction.
Therefore, we first investigate different DA integration methods for video classification and show that: 1) aligning the features that encode temporal dynamics outperforms aligning only spatial features. 2) to effectively align domains spatio-temporally, \textit{which features} to align is more important than \textit{what DA approaches} to use. To support our claims, we then propose \textit{Temporal Adversarial Adaptation Network (TA$^2$N)}, which simultaneously aligns and learns temporal dynamics, outperforming other approaches which naively apply more sophisticated image-based DA methods for videos. 

The temporal dynamics in videos can be represented as a combination of multiple local temporal features corresponding to different motion characteristics. Not all of the local temporal features equally contribute to the overall domain shift. We want to focus more on aligning those which have high contribution to the overall domain shift, such as the local temporal features connected by thicker green arrows shown in \Cref{fig:overview_video_DA}.
Therefore, we propose \textbf{Temporal Attentive Adversarial Adaptation Network (TA$\boldsymbol{^3}$N)} to explicitly attend to the temporal dynamics by taking into account the domain distribution discrepancy.
In this way, the temporal dynamics which contribute more to the overall domain shift will be focused on, leading to more effective temporal alignment. TA$^3$N achieves state-of-the-art performance on all four investigated video DA datasets.

In summary, our contributions are three-fold:
\begin{enumerate}
    \item \textbf{Video DA Dataset Collection}: 
    We collect two large-scale video DA datasets, \textit{UCF-HMDB$_{full}$} and \textit{Kinetics-Gameplay}, to investigate the domain discrepancy problem across videos, which is an under-explored research problem.
    To our knowledge, they are by far the largest datasets for video DA problems.
    
    \item \textbf{Feature Alignment Exploration for Video DA}:
    We investigate different DA integration approaches for videos and provide a strategy to effectively align domains spatio-temporally for videos by aligning temporal relation features. We propose this simple but effective approach, \textit{TA$^2$N}, to demonstrate the importance of determining \textit{what} to align over the DA method to use.
    
    \item \textbf{Temporal Attentive Adversarial Adaptation Network (TA$\boldsymbol{^3}$N)}: 
    We propose \textit{TA$^3$N}, which simultaneously aligns domains, encodes temporal dynamics into video representations, and attends to representations with domain distribution discrepancy. TA$^3$N achieves state-of-the-art performance on both small- and large-scale cross-domain video datasets. 

\end{enumerate}

\section{Related Works} 
\textbf{Video Classification.}
With the rise of deep convolutional neural networks (CNNs), recent work for video classification mainly aims to learn compact spatio-temporal representations by leveraging CNNs for spatial information and designing various architectures to exploit temporal dynamics~\cite{karpathy2014large}. In addition to separating spatial and temporal learning, some works propose different architectures to encode spatio-temporal representations with consideration of the trade-off between performance and computational cost~\cite{tran2015learning, carreira2017quo, qiu2017learning, tran2018closer}. Another branch of work utilizes optical flow to compensate for the lack of temporal information in raw RGB frames~\cite{simonyan2014two, feichtenhofer2016convolutional, wang2016temporal, carreira2017quo, ma2018ts}. Moreover, some works extract temporal dependencies between frames for video tasks by utilizing recurrent neural networks (RNNs)~\cite{donahue2015long}, attention~\cite{long2018attention, ma2018attend} and relation modules~\cite{zhou2018temporalrelation}.
Note that we focus on attending to the temporal dynamics to effectively align domains and we consider other modalities, e.g. optical flow, to be complementary to our method.

\textbf{Domain Adaptation.}
Most recent DA approaches are based on deep learning architectures designed for addressing the domain shift problems given the fact that the deep CNN features without any DA method outperform traditional DA methods using hand-crafted features~\cite{donahue2014decaf}. Most DA approaches follow the two-branch (source and target) architecture, and aim to find a common feature space between the source and target domains. The models are therefore optimized with a combination of \textit{classification} and \textit{domain} losses~\cite{csurka2017comprehensive}. 

One of the main classes of methods used is \textit{Discrepancy-based DA}, whose metrics are designed to measure the distance between source and target feature distributions, including variations of maximum mean discrepancy (MMD)~\cite{long2015learning, long2016unsupervised, zellinger2017central, yan2017mind, long2017deep} and the CORAL function~\cite{sun2016deep}. By diminishing the distance of distributions, discrepancy-based DA methods reduce the gap across domains. 
Another common method, \textit{Adversarial-based DA}, adopts a similar concept as GANs~\cite{goodfellow2014generative} by integrating domain discriminators into the architectures. Through the adversarial objectives, the discriminators are optimized to classify different domains, while the feature extractors are optimized in the opposite direction. ADDA~\cite{tzeng2017adversarial} uses an inverted label GAN loss to split the optimization into two parts: one for the discriminator and the other for the generator. In contrast, the gradient reversal layer (GRL) is used in some work~\cite{ganin2015unsupervised, ganin2016domain, zhang2018collaborative} to invert the gradients so that the discriminator and generator are optimized simultaneously. Additionally, \textit{Normalization-based DA}~\cite{li2017revisiting, li2018adaptive} adapts batch normalization~\cite{ioffe2015batch} to DA problems by calculating two separate statistics, representing source and target, for normalization. Furthermore, \textit{Ensemble-based DA}~\cite{french2018self, saito2018adversarial, saito2018maximum, lee2019sliced} builds a target branch ensemble by incorporating multiple target branches. Recently, TADA~\cite{wang2019transferable} adopts the attention mechanism to adapt the transferable regions. We extend these concepts to spatio-temporal domains, aiming to attend to the important parts of temporal dynamics for alignment.

\textbf{Video Domain Adaptation.}
Unlike image-based DA, video-based DA is still an under-explored area. Only a few works focus on small-scale video DA with only a few overlapping categories~\cite{sultani2014human, xu2016dual, jamal2018deep}. \cite{sultani2014human} improves the domain generalizability by decreasing the effect of the background. \cite{xu2016dual} maps source and target features to a common feature space using shallow neural networks. AMLS~\cite{jamal2018deep} adapts pre-extracted C3D~\cite{tran2015learning} features on a Grassmann manifold obtained using PCA. However, the datasets used in the above works are too small to have enough domain shift to evaluate DA performance. Therefore, we propose two larger cross-domain datasets \textit{UCF-HMDB$_{full}$} and \textit{Kinetics-Gameplay}, and provide benchmarks with different baseline approaches. 
Recently, TSRNet~\cite{zhang2019learning} transfers knowledge for action localization using MMD, but only aligns the video-level features. Instead, our \textit{TA$^3$N} simultaneously attends, aligns, and encodes temporal dynamics into video features.

\section{Technical Approach}
We first introduce our baseline model which simply extends image-base DA for videos using the temporal pooling mechanism (\Cref{sec:baseline}). And then we investigate better ways to incorporate temporal dynamics for video DA (\Cref{sec:TAAN}), and describe our final proposed method with the domain attention mechanism (\Cref{sec:TAAAN}).

\subsection{Baseline Model} \label{sec:baseline}
\begin{figure}[!ht]
\centering
\includegraphics[scale=0.373]{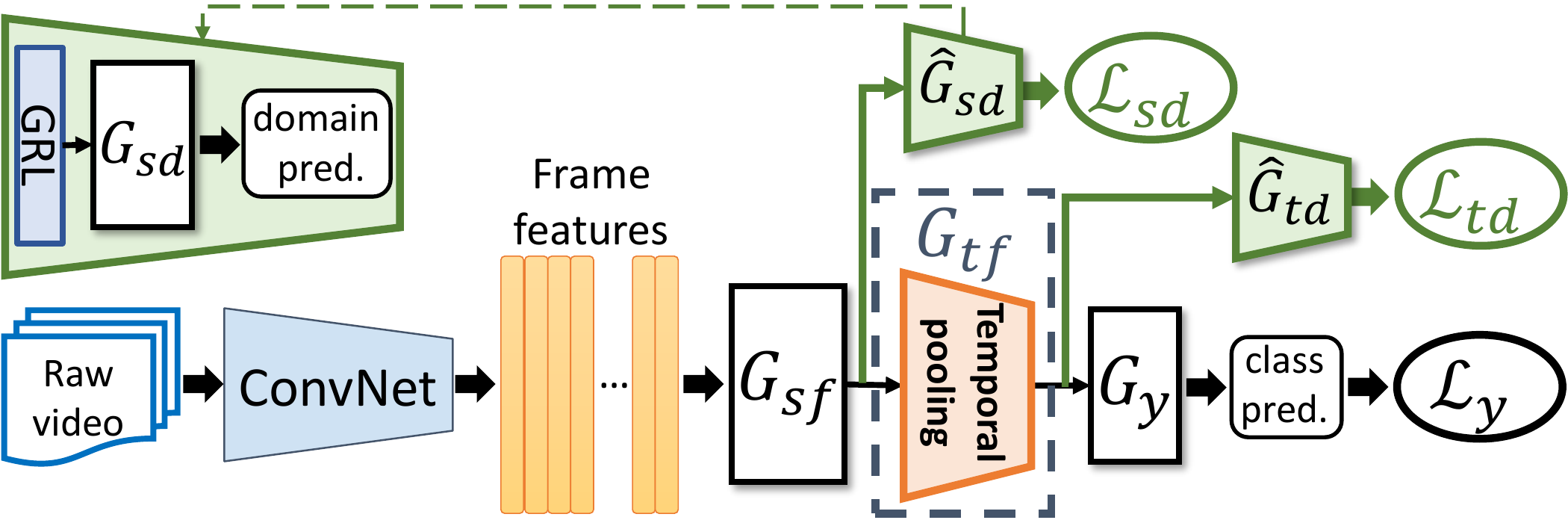}
\caption{Baseline architecture (TemPooling) with the adversarial discriminators $\hat{G}_{sd}$ and $\hat{G}_{td}$. $\mathcal{L}_y$ is the class prediction loss, and $\mathcal{L}_{sd}$ and $\mathcal{L}_{td}$ are the domain losses. See the detailed architecture in the supplementary material.}
\label{fig:TemPooling_RevGrad}
\end{figure}

Given the recent success of large-scale video classification using CNNs~\cite{karpathy2014large}, we build our baseline on such architectures, as shown in the lower part of Figure~\ref{fig:TemPooling_RevGrad}. 

We first feed the input video $X_i=\{x_i^1, x_i^2, ..., x_i^K\}$ extracted from ResNet~\cite{he2016deep} pre-trained on ImageNet into our model, where $x^j_i$ is the $j$th frame-level feature representation of the $i$th video. The model can be divided into two parts: 1) \textit{Spatial module $G_{sf}(.;\theta_{sf})$}, which consists of multilayer perceptrons (MLP) that aims to convert the general-purpose feature vectors into task-driven feature vectors, where the task is video classification in this paper; 
2) \textit{Temporal module $G_{tf}(.;\theta_{tf})$} aggregates the frame-level feature vectors to form a single video-level feature vector for each video. In our baseline architecture, we conduct mean-pooling along the temporal direction to generate video-level feature vectors, and note it as \textit{TemPooling}.
Finally, another fully-connected layer $G_y(.;\theta_y)$ converts the video-level features into the final predictions, which are used to calculate the class prediction loss $\mathcal{L}_y$. 

Similar to image-based DA problems, the baseline approach is not able to generalize to data from different domains due to domain shift. 
Therefore, we integrate TemPooling with the unsupervised DA method inspired by one of the most popular adversarial-based approaches, DANN~\cite{ganin2015unsupervised, ganin2016domain}. 
The main idea is to add additional 
domain classifiers $G_d(.;\theta_d)$, to discriminate whether the data is from the source or target domain. 
Before back-propagating the gradients to the main model, a gradient reversal layer (GRL) is inserted between $G_d$ and the main model to invert the gradient, as shown in Figure~\ref{fig:TemPooling_RevGrad}. 
During adversarial training, the parameters $\theta_{sf}$ are learned by maximizing the domain discrimination loss $\mathcal{L}_d$, and parameters $\theta_d$ are learned by minimizing $\mathcal{L}_d$ with the domain label $d$. 
Therefore, the feature generator $G_f$ will be optimized to gradually align the feature distributions between the two domains. 

In this paper, we note the \textit{Adversarial Discriminator $\hat{G}_d$} as the combination of a gradient reversal layer (GRL) and a domain classifier, and insert $\hat{G}_d$ into TemPooling in two ways: 1) $\hat{G}_{sd}$: show how directly applying image-based DA approaches can benefit video DA; 2) $\hat{G}_{td}$: indicate how DA on temporal-dynamics-encoded features benefits video DA. 

The prediction loss $\mathcal{L}_y$, spatial domain loss $\mathcal{L}_{sd}$ and temporal domain loss $\mathcal{L}_{td}$ can be expressed as follows (ignoring all the parameter symbols through the paper to save space):
\begin{equation} \label{eq:loss-pred}
\small
\mathcal{L}^i_y = L_y(G_y(G_{tf}(G_{sf}(X_i))),y_i)
\end{equation}
\begin{equation} \label{eq:loss-domain-spatial}
\small
\mathcal{L}^i_{sd} = \frac{1}{K}\sum_{j=1}^{K}L_d(G_{sd}(G_{sf}(x^j_i)),d_i)
\end{equation}
\begin{equation} \label{eq:loss-domain-temporal}
\small
\mathcal{L}^i_{td} = L_d(G_{td}(G_{tf}(G_{sf}(X_i))),d_i)
\end{equation}
where $K$ is the number of frames sampled from each video. \textit{L} is the cross entropy loss function.

The overall loss can be expressed as follows:
\begin{equation} \label{eq:loss-all-baseline}
\small
\begin{split}
\mathcal{L} = \frac{1}{N_S}\sum_{i=1}^{N_S}\mathcal{L}^i_y 
-\frac{1}{N_{S\cup T}}\sum_{i=1}^{N_{S\cup T}}(\lambda_s\mathcal{L}^i_{sd}+\lambda_t\mathcal{L}^i_{td})
\end{split}
\end{equation}
where $N_S$ equals the number of source data, and $N_{S\cup T}$ equals the number of all data. $\lambda_s$ and $\lambda_t$ is the trade-off weighting for spatial and temporal domain loss.

\subsection{Integration of Temporal Dynamics with DA} \label{sec:TAAN}
One main drawback of directly integrating image-based DA approaches into our baseline architecture is that the feature representations learned in the model are mainly from the spatial features. Although we implicitly encode the temporal information by the temporal pooling mechanism, the relation between frames is still missing. 
Therefore, we would like to address two questions: 1) \textit{Does the video DA problem benefit from encoding temporal dynamics into features?} 2) \textit{Instead of only modifying feature encoding methods, how can DA be further integrated while encoding temporal dynamics into features?}

To answer the first question, given the fact that humans can recognize actions by reasoning the observations across time, we propose the \textit{TemRelation} architecture by replacing the temporal pooling mechanism with the Temporal Relation module, which is modified from \cite{santoro2017simple, zhou2018temporalrelation}, as shown in Figure~\ref{fig:TemRelation_RevGrad_TransAttn}. 

The $n$-frame temporal relation is defined by the function:
\begin{equation} \label{eq:temporal-relation}
\small
R_n(V_i)=\sum_{m}g_{\phi^{(n)}}((V^n_i)_m)
\end{equation}
where $(V^n_i)_m=\{v_i^a, v_i^b, ...\}_m$ is the $m$th set of frame-level representations from $n$ temporal-ordered sampled frames. $a$ and $b$ are the frame indices.
We fuse the feature vectors that are time-ordered with the function $g_{\phi^{(n)}}$, which is an MLP with parameters $\phi^{(n)}$. 
To capture temporal relations at multiple time scales, we sum up all the $n$-frame relation features into the final video representation. 
In this way, the temporal dynamics are explicitly encoded into features. We then insert $\hat{G}_d$ into TemRelation as we did for TemPooling. 

Although aligning temporal-dynamic-encoded features benefits video DA, feature encoding and DA are still two separate processes, leading to sub-optimal DA performance. Therefore, we address the second question by proposing \textbf{Temporal Adversarial Adaptation Network (TA$^2$N)}, which explicitly integrates $\hat{G}_d$ \textit{inside} the Temporal module to align the model across domains while learning temporal dynamics. 
Specifically, we integrate each $n$-frame relation with a corresponding relation discriminator $\hat{G}^n_{rd}$ because different $n$-frame relations represent different temporal characteristics, which correspond to different parts of actions.
The relation domain loss $\mathcal{L}_{rd}$ can be expressed as follows:
\begin{equation} \label{eq:loss-domain-relation}
\small
\mathcal{L}^i_{rd} = \frac{1}{K-1}\sum_{n=2}^{K}L_d(G^n_{rd}(R_n(G_{sf}(X_i))),d_i)
\end{equation}
The experimental results show that our integration strategy can effectively align domains spatio-temporally for videos, and outperform those which are extended from sophisticated DA approaches although TA$^2$N is adopted from a simpler DA method (DANN) (see details in \Cref{table:sota_ucf-hmdb_full,table:sota_hmdb-ucf_full,table:sota_kinetics-gameplay}).

\subsection{Temporal Attentive Alignment for Videos} \label{sec:TAAAN}
\begin{figure}[!t]
\centering
\includegraphics[width=0.475\textwidth]{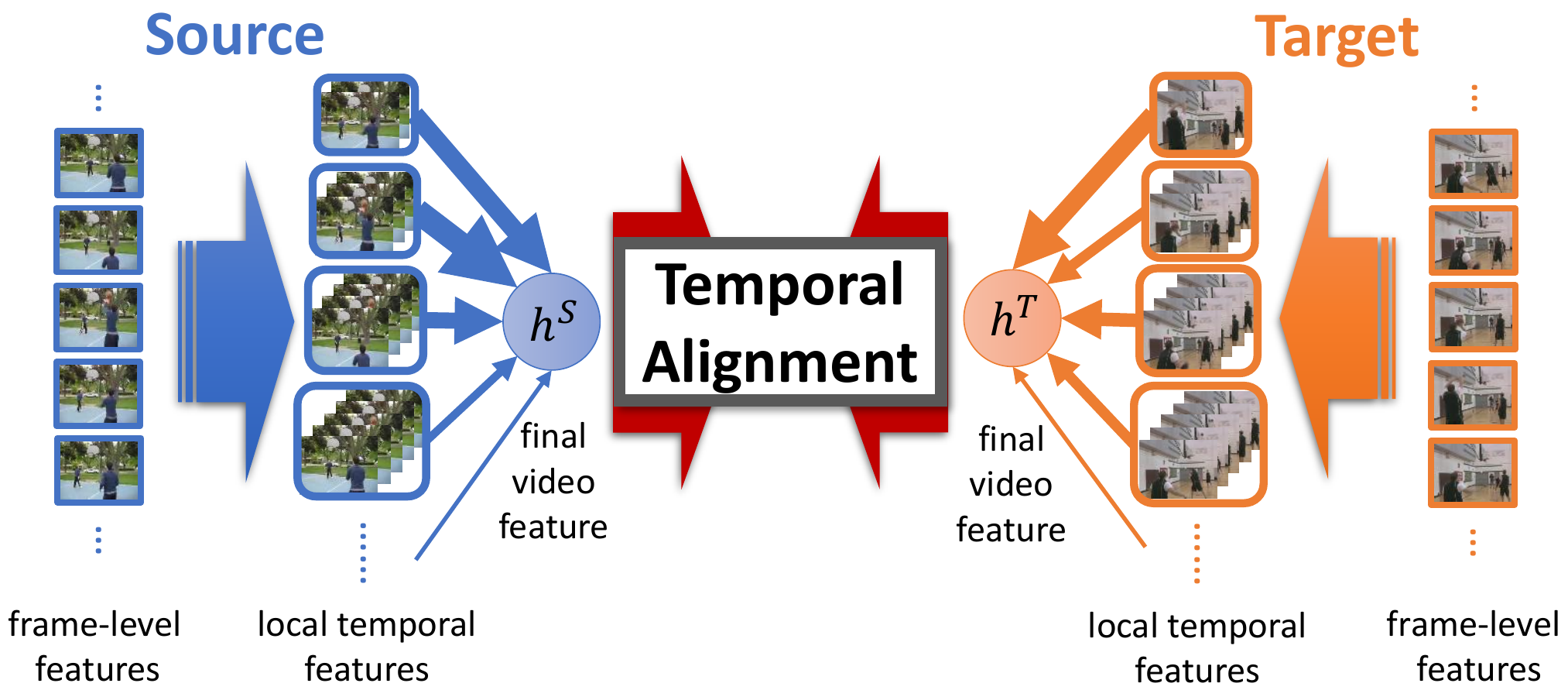}
\caption{The domain attention mechanism in TA$^3$N.
Thicker arrows correspond to larger attention weights.
}
\label{fig:overview_domain_attention}
\end{figure}

The final video representation of TA$^2$N is generated by aggregating multiple local temporal features. Although aligning temporal features across domains benefits video DA, not all the features are equally important to align. In order to effectively align overall temporal dynamics, we want to focus more on aligning the local temporal features which have larger domain discrepancy. 
Therefore, we represent the final video representation as a combination of local temporal features with different attention weighting, as shown in \Cref{fig:overview_domain_attention}, and aim to attend to features of interest that are domain discriminative so that the DA mechanism can focus on aligning those features. 
The main question becomes: \textit{How to incorporate domain discrepancy for attention?} 

\begin{figure*}[!t]
\centering
\includegraphics[width=\textwidth]{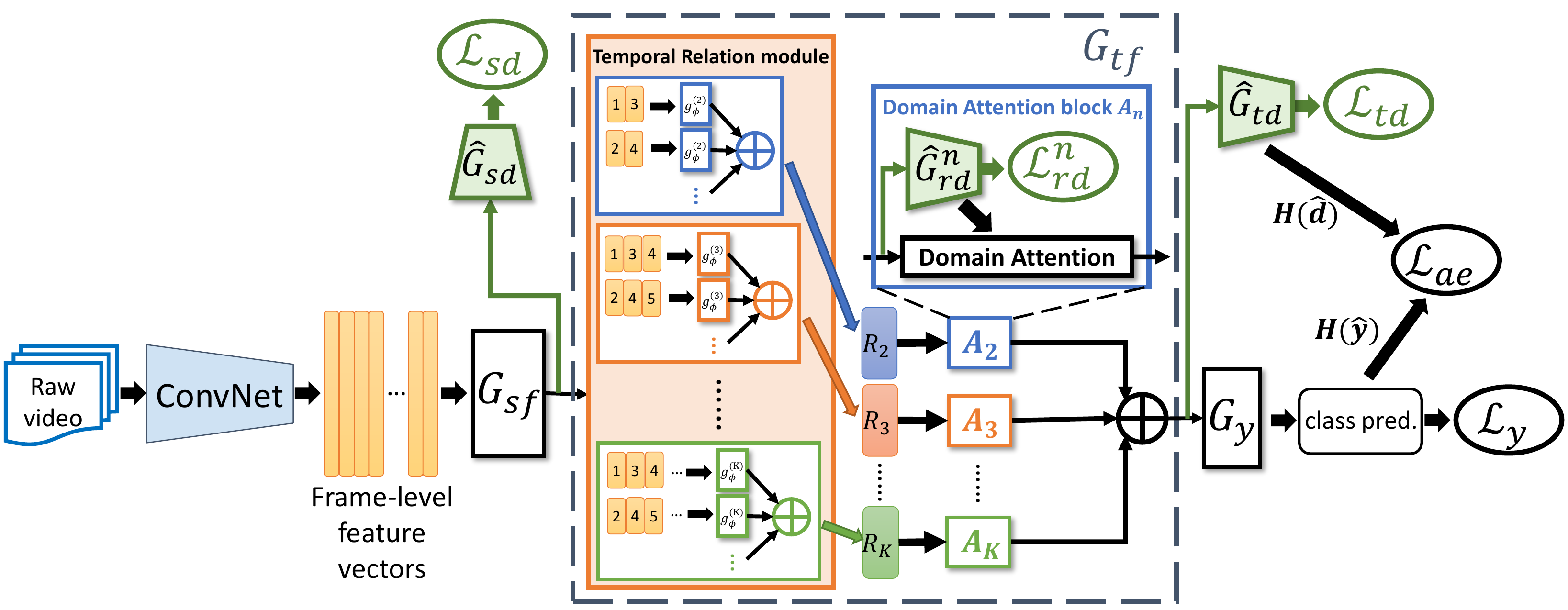}
\caption{The overall architecture of the proposed Temporal Attentive Adversarial Adaptation Network (TA$^3$N). In the temporal relation module, time-ordered frames are used to generate $K$-1 relation feature representations $\textit{\textbf{R}}=\{R_2, ..., R_K\}$, where $R_n$ corresponds to the $n$-frame relation (the numbers in this figure are examples of time indices). 
After attending with the domain predictions from relation discriminators $G^n_{rd}$, the relation features are summed up to the final video representation. The attentive entropy loss $\mathcal{L}_{ae}$, which is calculated by domain entropy $H(\hat{d})$ and class entropy $H(\hat{y})$, aims to enhance the certainty of those videos that are more similar across domains. See the detailed architecture in the supplementary material.
}
\label{fig:TemRelation_RevGrad_TransAttn}
\end{figure*}

To address this, we propose \textbf{Temporal Attentive Adversarial Adaptation Network (TA$^3$N)}, as shown in \Cref{fig:TemRelation_RevGrad_TransAttn}, by introducing the \textit{domain attention} mechanism, which utilize the entropy criterion to generate the domain attention value for each $n$-frame relation feature as below:
\begin{equation} \label{eq:attention-weight}
\small
w^n_i = 1 - H(\hat{d}^n_i)
\end{equation}
where $\hat{d}^n_i$ is the output of $G^n_{rd}$ for the $i$th video. $H(p) = -\sum_{k}p_k\cdot\log(p_k)$ is the entropy function to measure uncertainty. $w^n_i$ increases when $H(\hat{d}^n_i)$ decreases, which means the domains can be distinguished well.
We also add a residual connection for more stable optimization. Therefore, the final video feature representation $h_i$ generated from attended local temporal features, which are learned by local temporal modules $G^{(n)}_{tf}$, can be expressed as:
\begin{equation} \label{eq:attention-residual}
\small
h_i = \sum_{n=2}^{K}(w^n_i + 1) \cdot G^{(n)}_{tf}(G_{sf}(X_i))
\end{equation}

Finally, we add the minimum entropy regularization to refine the classifier adaptation. However, we only want to minimize the entropy for the videos that are similar across domains. 
Therefore, we attend to the videos which have low domain discrepancy, so that we can focus more on minimizing the entropy for these videos. 
The attentive entropy loss $\mathcal{L}_{ae}$ can be expressed as follows:
\begin{equation} \label{eq:loss-attentive-entropy}
\small
\mathcal{L}^i_{ae}=(1+H(\hat{d}_i)) \cdot H(\hat{y}_i)
\end{equation}
where $\hat{d}_i$ and $\hat{y}_i$ is the output of $G_{td}$ and $G_y$, respectively. We also adopt the residual connection for stability.

By combining \Cref{eq:loss-pred,eq:loss-domain-spatial,eq:loss-domain-temporal,eq:loss-domain-relation,eq:loss-attentive-entropy}, and replacing $G_{sf}$ and $G_{tf}$ with $h_i$ by \Cref{eq:attention-residual}, 
the overall loss of TA$^3$N can be expressed as follows:
\begin{equation} \label{eq:loss-all}
\small
\begin{split}
&\mathcal{L} = \frac{1}{N_S}\sum_{i=1}^{N_S}\mathcal{L}^i_y + \frac{1}{N_{S\cup T}}\sum_{i=1}^{N_{S\cup T}}\gamma\mathcal{L}^i_{ae} \\
&-\frac{1}{N_{S\cup T}}\sum_{i=1}^{N_{S\cup T}}(\lambda^s\mathcal{L}^{i}_{sd}+\lambda^r\mathcal{L}^{i}_{rd}+\lambda^t\mathcal{L}^{i}_{td})
\end{split}
\end{equation}
where $\lambda^s$, $\lambda^r$ and $\lambda^t$ is the trade-off weighting for each domain loss. $\gamma$ is the weighting for the attentive entropy loss. All the weightings are chosen via grid search. 

Our proposed TA$^3$N and TADA~\cite{wang2019transferable} both utilize entropy functions for attention but with different perspectives. TADA aims to focus on the foreground objects for image DA, while TA$^3$N aims to find important and discriminative parts of temporal dynamics to align for video DA.

\section{Datasets}
There are very few benchmark datasets for video DA, and only small-scale datasets have been widely used~\cite{sultani2014human, xu2016dual, jamal2018deep}.
Therefore, we specifically create two cross-domain datasets to evaluate the proposed approaches for the video DA problem, as shown in \Cref{table:dataset}. 
For more details about the datasets, please refer to the supplementary material. 

\begin{table*}[!t]
\centering
    \begin{tabular}{c|c|c|c|c}
     & UCF-HMDB$_{small}$ & UCF-Olympic & UCF-HMDB$_{full}$ & Kinetics-Gameplay \\ \hline
    length (sec.) & 1 - 21 & 1 - 39 & 1 - 33 & 1 - 10 \\ \hline
    class \# & 5 & 6 & 12 & 30 \\ \hline
    video \# & 1171 & 1145 & 3209 & 49998 \\ \hline
    \end{tabular}
\caption{The comparison of the cross-domain video datasets.}
\label{table:dataset}
\end{table*}

\subsection{UCF-HMDB$\boldsymbol{_{full}}$}
We extend UCF-HMDB$_{small}$~\cite{sultani2014human}, which only selects 5 visually highly similar categories, by collecting all of the relevant and overlapping categories between UCF101~\cite{soomro2012ucf101} and HMDB51~\cite{kuehne2011hmdb}, which results in 12 categories.
We follow the official split method to separate training and validation sets. This dataset, \textbf{UCF-HMDB$\boldsymbol{_{full}}$}, includes more than 3000 video clips, which is around 3 times larger than UCF-HMDB$_{small}$ and UCF-Olympic.

\subsection{Kinetics-Gameplay}
In addition to real-world videos, we are also interested in virtual-world videos for DA. While there are more than ten real-world video datasets, there is a limited number of virtual-world datasets for video classification. 
It is mainly because rendering realistic human actions using game engines requires gaming graphics expertise which is time-consuming. 
Therefore, we create the \textit{Gameplay} dataset by collecting gameplay videos from currently popular video games, \textit{Detroit: Become Human} and \textit{Fortnite}, to build our own video dataset for the virtual domain. 
For the real domain, we use one of the largest public video datasets \textit{Kinetics-600}~\cite{kay2017kinetics, carreira2018kinetics}. 
We follow the closed-set DA setting~\cite{peng2018syn2real} to select 30 overlapping categories between the Kinetics-600 and Gameplay datasets
to build the \textbf{Kinetics-Gameplay} dataset with both domains, including around 50K video clips.
See the supplementary material for the complete statistics and example snapshots.

\section{Experiments}
We therefore evaluate DA approaches on four datasets: UCF-Olympic, UCF-HMDB$_{small}$, UCF-HMDB$_{full}$ and Kinetics-Gameplay.

\subsection{Experimental Setup}
\textbf{UCF-Olympic} and \textbf{UCF-HMDB$\boldsymbol{_{small}}$}.
First, we evaluate our approaches on UCF-Olympic and UCF-HMDB$_{small}$, and compare with all other works that also evaluate on these two datasets~\cite{sultani2014human, xu2016dual, jamal2018deep}. 
We follow the default settings, but the method to split the UCF video clips into training and validations sets is not specified in these papers, so we follow the official split method from UCF101~\cite{soomro2012ucf101}.

\textbf{UCF-HMDB$\boldsymbol{_{full}}$} and \textbf{Kinetics-Gameplay}.
For the self-collected datasets, we follow the common experimental protocol of unsupervised DA~\cite{peng2018syn2real}: the training data consists of labeled data from the source domain and unlabeled data from the target domain, and the validation data is all from the target domain. However, unlike most of the image DA settings, our training and validation data in both domains are separate to avoid potentially overfitting while aligning different domains.
To compare with image-based DA approaches, we extend several state-of-the-art methods~~\cite{ganin2016domain, long2017deep, li2018adaptive, saito2018maximum} for video DA with our TemPooling and TemRelation architectures, as shown in \Cref{table:sota_ucf-hmdb_full,table:sota_hmdb-ucf_full,table:sota_kinetics-gameplay}. 
The difference between the ``Target only" and ``Source only" settings is the domain used for training. The ``Target only" setting can be regarded as the upper bound without domain shift while the ``Source only" setting shows the lower bound which directly applies the model trained with source data to the target domain without modification. 
See supplementary materials for full implementation details.

\subsection{Experimental Results} \label{sec:experimental_results}
\textbf{UCF-Olympic} and \textbf{UCF-HMDB$\boldsymbol{_{small}}$}.
In these two datasets, our approach outperforms all the previous methods by at least 6.5\% absolute difference (98.15\% - 91.60\%) on the ``U $\rightarrow$ O" setting, and 9\% difference (99.33\% - 90.25\%) on the ``U $\rightarrow$ H" setting, as shown in Table~\ref{table:sota_ucf-olympic_ucf-hmdb-small}.

These results also show that the performance on these datasets is saturated. With a strong CNN as the backbone architecture, even our baseline architecture TemPooling can achieve high accuracy without any DA method (e.g. 96.3\% for ``U $\rightarrow$ O").
This suggests that these two datasets are not enough to evaluate more sophisticated DA approaches, so larger-scale datasets for video DA are needed.

\begin{table}[!t]
\centering
\footnotesize
    \begin{tabular}{c|c|c|c|c}
    Source $\rightarrow$ Target & U $\rightarrow$ O & O $\rightarrow$ U & U $\rightarrow$ H & H $\rightarrow$ U \\ \hline
    W. Sultani et al.~\cite{sultani2014human} & 33.33 & 47.91 & 68.70 & 68.67 \\ 
    T. Xu et al. ~\cite{xu2016dual} & 87.00 & 75.00 & 82.00 & 82.00 \\ 
    AMLS (GFK)~\cite{jamal2018deep}$\dag$ & 84.65 & 86.44 & 89.53 & 95.36 \\ 
    AMLS (SA)~\cite{jamal2018deep}$\dag$ & 83.92 & 86.07 & 90.25 & 94.40 \\ 
    DAAA~\cite{jamal2018deep}$\dag\ddag$ & 91.60 & 89.96 & - & - \\ \hline
    TemPooling & 96.30 & 87.08 & 98.67 & 97.35 \\ 
    TemPooling + DANN~\cite{ganin2016domain} & \textbf{98.15} & 90.00 & \textbf{99.33} & 98.41 \\ \hline\hline 
    Ours (TA$^2$N) & \textbf{98.15} & 91.67 & \textbf{99.33} & \textbf{99.47} \\
    Ours (TA$^3$N) & \textbf{98.15} & \textbf{92.92} & \textbf{99.33} & \textbf{99.47} \\ \hline
    \end{tabular}
\caption{The accuracy (\%) for the state-of-the-art work on UCF-Olympic and UCF-HMDB$_{small}$ (U: UCF, O: Olympic, H: HMDB). $\dag$We only show their results which are fine-tuned with source data for fair comparison. Please refer to the supplementary material for more details.
$\ddag$\cite{jamal2018deep} did not test DAAA on UCF-HMDB$_{small}$.
} 
\label{table:sota_ucf-olympic_ucf-hmdb-small}
\end{table}

\textbf{UCF-HMDB$\boldsymbol{_{full}}$}.
We then evaluate our approaches and compare with other image-based DA approaches on the UCF-HMDB$_{full}$ dataset, as shown in \Cref{table:sota_ucf-hmdb_full,table:sota_hmdb-ucf_full}. The accuracy difference between ``Target only" and ``Source only" indicates the \textit{domain gap}. 
The gaps for the HMDB dataset are 11.11\% for TemRelation and 10.28\% for TemPooling (see \Cref{table:sota_ucf-hmdb_full}), and the gaps for the UCF dataset are 21.01\% for TemRelation and 17.16\% for TemPooling (see \Cref{table:sota_hmdb-ucf_full}). It is worth noting that the ``Source only" accuracy of our baseline architecture (TemPooling) on UCF-HMDB$_{full}$ is much lower than UCF-HMDB$_{small}$ (e.g. 28.39 lower for ``U $\rightarrow$ H"), which implies that UCF-HMDB$_{full}$ contains much larger domain discrepancy than UCF-HMDB$_{small}$.
The value ``Gain" is the difference from the ``Source only" accuracy, which directly indicates the effectiveness of the DA approaches. 
We now answer the two questions for video DA in Section~\ref{sec:TAAN} (see \Cref{table:sota_ucf-hmdb_full,table:sota_hmdb-ucf_full}):
\begin{enumerate}
    \item \textit{Does the video DA problem benefit from encoding temporal dynamics into features?}

    From \Cref{table:sota_ucf-hmdb_full,table:sota_hmdb-ucf_full}, we see that for the same DA method, TemRelation outperforms TemPooling in most cases, especially for the gain value. For example, 
    ``TemPooling+DANN" reaches 0.83\% absolute accuracy gain on the ``U $\rightarrow$ H" setting and 0.17\% gain on the ``H $\rightarrow$ U" setting while ``TemRelation+DANN" reaches 3.61\% gain on ``U $\rightarrow$ H" and 2.45\% gain on ``H $\rightarrow$ U".
    This means that applying DA approaches to the video representations which encode the temporal dynamics improves the overall performance for cross-domain video classification.

    \item \textit{How to further integrate DA while encoding temporal dynamics into features?} 
    
    Although integrating TemRelation with image-based DA approaches generally has better alignment performance than the baseline (TemPooling), feature encoding and DA are still two separate processes. The alignment happens only before and after the temporal dynamics are encoded in features. In order to explicitly force alignment of the temporal dynamics across domains, we propose TA$^2$N, which reaches 77.22\% (5.55\% gain) on ``U $\rightarrow$ H" and 80.56\% (6.66\% gain) on ``H $\rightarrow$ U". \Cref{table:sota_ucf-hmdb_full,table:sota_hmdb-ucf_full} show that although TA$^2$N is adopted from a simple DA method (DANN), it still outperforms other approaches which are extended from more sophisticated DA methods but do not follow our strategy.
\end{enumerate}

Finally, with the domain attention mechanism, our proposed \textbf{TA$\boldsymbol{^3}$N} reaches 78.33\%  (6.66\% gain) on ``U $\rightarrow$ H" and 81.79\% (7.88\% gain) on ``H $\rightarrow$ U", achieving state-of-the-art performance on UCF-HMDB$_{full}$ in terms of accuracy and gain, as shown in \Cref{table:sota_ucf-hmdb_full,table:sota_hmdb-ucf_full}.

\begin{table}[!t]
\centering
\small
    \begin{tabular}{c|c|c|c|c}
    Temporal Module & \multicolumn{2}{c|}{TemPooling} & \multicolumn{2}{c}{TemRelation} \\ \hline
     & Acc. & Gain & Acc. & Gain \\ \hline
    Target only & 80.56 & - & 82.78 & - \\ \hline
    Source only & 70.28 & - & 71.67 & - \\ 
    DANN~\cite{ganin2016domain} & 71.11 & 0.83 & 75.28 & 3.61 \\ 
    JAN~\cite{long2017deep} & 71.39 & 1.11 & 74.72 & 3.05 \\ 
    AdaBN~\cite{li2018adaptive} & 75.56 & 5.28 & 72.22 & 0.55 \\ 
    MCD~\cite{saito2018maximum} & 71.67 & 1.39 & 73.89 & 2.22 \\ \hline
    Ours (TA$^2$N) & N/A & - & 77.22 & 5.55 \\ 
    Ours (TA$^3$N) & N/A & - & \textbf{78.33} & \textbf{6.66} \\ \hline
    \end{tabular}
\caption{The comparison of accuracy (\%) with other approaches on UCF-HMDB$_{full}$ (U $\rightarrow$ H). Gain represents the absolute difference from the ``Source only" accuracy. TA$^2$N and TA$^3$N are based on the TemRelation architecture, so they are not applicable to TemPooling.}
\label{table:sota_ucf-hmdb_full}
\end{table}

\begin{table}[!t]
\centering
\small
    \begin{tabular}{c|c|c|c|c}
    Temporal Module & \multicolumn{2}{c|}{TemPooling} & \multicolumn{2}{c}{TemRelation} \\ \hline
     & Acc. & Gain & Acc. & Gain \\ \hline
    Target only & 92.12 & - & 94.92 & - \\ \hline
    Source only & 74.96 & - & 73.91 & - \\ 
    DANN~\cite{ganin2016domain} & 75.13 & 0.17 & 76.36 & 2.45 \\ 
    JAN~\cite{long2017deep} & 80.04 & 5.08 & 79.69 & 5.79 \\ 
    AdaBN~\cite{li2018adaptive} & 76.36 & 1.40 & 77.41 & 3.51 \\ 
    MCD~\cite{saito2018maximum} & 76.18 & 1.23 & 79.34 & 5.44 \\ \hline
    Ours (TA$^2$N) & N/A & - & 80.56 & 6.66 \\
    Ours (TA$^3$N) & N/A & - & \textbf{81.79} & \textbf{7.88} \\ \hline
    \end{tabular}
\caption{The comparison of accuracy (\%) with other approaches on UCF-HMDB$_{full}$ (H $\rightarrow$ U).}
\label{table:sota_hmdb-ucf_full}
\end{table}

\textbf{Kinetics-Gameplay}.
Kinetics-Gameplay is much more challenging than UCF-HMDB$_{full}$ because the data is from real and virtual domains, which have more severe domain shifts. Here we only utilize TemRelation as our backbone architecture since it is proved to outperform TemPooling on UCF-HMDB$_{full}$. Table~\ref{table:sota_kinetics-gameplay} shows that the accuracy gap between ``Source only" and ``Target only" is 47.27\%, which is more than twice the number in UCF-HMDB$_{full}$. In this dataset, TA$^3$N also outperforms all the other DA approaches by increasing the ``Source only " accuracy from 17.22\% to 27.50\%.

\begin{table}[!t]
\centering
    \begin{tabular}{c|c|c}
     & Acc. & Gain \\ \hline
    Target only & 64.49 & - \\ \hline
    Source only & 17.22 & - \\ 
    DANN~\cite{ganin2016domain} & 20.56 & 3.34 \\ 
    JAN~\cite{long2017deep} & 18.16 & 0.94 \\ 
    AdaBN~\cite{li2018adaptive} & 20.29 & 3.07 \\ 
    MCD~\cite{saito2018maximum} & 19.76 & 2.54 \\ \hline
    Ours (TA$^2$N) & 24.30 & 7.08 \\
    Ours (TA$^3$N) & \textbf{27.50} & \textbf{10.28} \\ \hline
    \end{tabular}
\caption{The comparison of accuracy (\%) with other approaches on Kinetics-Gameplay. }
\label{table:sota_kinetics-gameplay}
\end{table}

\subsection{Ablation Study and Analysis}
\textbf{Integration of $\boldsymbol{\hat{G}_d}$}.
We use UCF-HMDB${_{full}}$ to investigate the performance for integrating $\hat{G}_d$ in different positions.
There are three ways to insert the adversarial discriminator into our architectures, where each corresponds to different feature representations, leading to three types of discriminators $\hat{G}_{sd}$, $\hat{G}_{td}$ and $\hat{G}_{rd}$, which are shown in Figure~\ref{fig:TemRelation_RevGrad_TransAttn} and the full experimental results are shown in Table~\ref{table:experiments_dann_position}. For the TemRelation architecture, the accuracy of utilizing $\hat{G}_{td}$ shows better performance than utilizing $\hat{G}_{sd}$ (averagely 0.58\% absolute gain improvement across two tasks), while the accuracies are the same for TemPooling. This means that the temporal relation module can encode temporal dynamics that help the video DA problem, but temporal pooling cannot. Utilizing the relation discriminator $\hat{G}_{rd}$ can further improve the performance (0.92\% improvement) since we simultaneously align and learn the temporal dynamics across domains. Finally, by combining all three discriminators, TA$^2$N improves even more (4.20\% improvement).

\begin{table}[!t]
\centering
\scriptsize
    \begin{tabular}{c|c|c|c|c}
    S $\rightarrow$ T & \multicolumn{2}{c|}{UCF $\rightarrow$ HMDB} & \multicolumn{2}{c}{HMDB $\rightarrow$ UCF} \\ \hline
    Temporal & \multirow{2}{*}{TemPooling} & \multirow{2}{*}{TemRelation} & \multirow{2}{*}{TemPooling} & \multirow{2}{*}{TemRelation} \\ 
    Module &  &  &  &  \\ \hline
    Target only & 80.56 (-) & 82.78 (-) & 92.12 (-) & 94.92 (-) \\ \hline
    Source only & 70.28 (-) & 71.67 (-) & 74.96 (-) & 73.91 (-) \\ 
    $\hat{G}_{sd}$ & 71.11 (0.83) & 74.44 (2.77) & 75.13 (0.17) & 74.44 (1.05) \\ 
    $\hat{G}_{td}$ & 71.11 (0.83) & 74.72 (3.05) & 75.13 (0.17) & 75.83 (1.93)  \\ 
    $\hat{G}_{rd}$ & - (-) & 76.11 (4.44) & - (-) & 75.13 (1.23) \\ \hline
    All $\hat{G}_d$ & 71.11 (0.83) & \textbf{77.22} (\textbf{5.55}) & 75.13 (0.17) & \textbf{80.56} (\textbf{6.66}) \\ \hline
    \end{tabular}
\caption{The full evaluation of accuracy (\%) for integrating $\hat{G}_d$ in different positions without the attention mechanism. Gain values are in (). }
\label{table:experiments_dann_position}
\end{table}

\textbf{Attention mechanism}.
In addition to TemRelation, we also apply the domain attention mechanism to TemPooling by attending to the raw frame features instead of relation features, and improve the performance as well, as shown in \Cref{table:attention-tempooling-temrelation}. 
This implies that video DA can benefit from the domain attention even if the backbone architecture does not encode temporal dynamics. We also compare the domain attention module with the general attention module, which calculates the attention weights via the \textit{FC-Tanh-FC-Softmax} architecture. However, it performs worse since the weights are computed within one domain, lacking of the consideration of domain discrepancy, as shown in \Cref{table:attention-comparison}.

\begin{table}[!t]
\centering
\scriptsize
    \begin{tabular}{c|c|c|c|c}
    S $\rightarrow$ T & \multicolumn{2}{c|}{UCF $\rightarrow$ HMDB} & \multicolumn{2}{c}{HMDB $\rightarrow$ UCF} \\ \hline
    Temporal & \multirow{2}{*}{TemPooling} & \multirow{2}{*}{TemRelation} & \multirow{2}{*}{TemPooling} & \multirow{2}{*}{TemRelation} \\ 
    Module &  &  &  &  \\ \hline
    Target only & 80.56 (-) & 82.78 (-) & 92.12 (-) & 94.92 (-) \\ \hline
    Source only & 70.28 (-) & 71.67 (-) & 74.96 (-) & 73.91 (-) \\ 
    All $\hat{G}_d$ & 71.11 (0.83) & 77.22 (5.55) & 75.13 (0.17) & 80.56 (6.66) \\ \hline
    All $\hat{G}_d$ & \multirow{2}{*}{\textbf{73.06 (2.78)}} & \multirow{2}{*}{\textbf{78.33 (6.66)}} & \multirow{2}{*}{\textbf{78.46 (3.50)}} & \multirow{2}{*}{\textbf{81.79 (7.88)}} \\ 
    +Domain Attn. &  &  &  &  \\ \hline
    \end{tabular}
\caption{The affect of the domain attention mechanism. }
\label{table:attention-tempooling-temrelation}
\end{table}

\begin{table}[!t]
\centering
\small
    \begin{tabular}{c|c|c}
    S $\rightarrow$ T & UCF $\rightarrow$ HMDB & HMDB $\rightarrow$ UCF \\ \hline
    Target only & 82.78 (-) & 94.92 (-) \\ \hline
    Source only & 71.67 (-) & 73.91 (-) \\ 
    No Attention & 77.22 (5.55) & 80.56 (6.66) \\ \hline
    General Attention & 77.22 (5.55) & 80.91 (7.00) \\ 
    Domain Attention & \textbf{78.33} (\textbf{6.66}) & \textbf{81.79} (\textbf{7.88}) \\ \hline
    \end{tabular}
\caption{The comparison of different attention methods.}
\label{table:attention-comparison}
\end{table}

\begin{figure}[!t]
  \begin{subfigure}[b]{0.235\textwidth}
    \includegraphics[width=\textwidth]{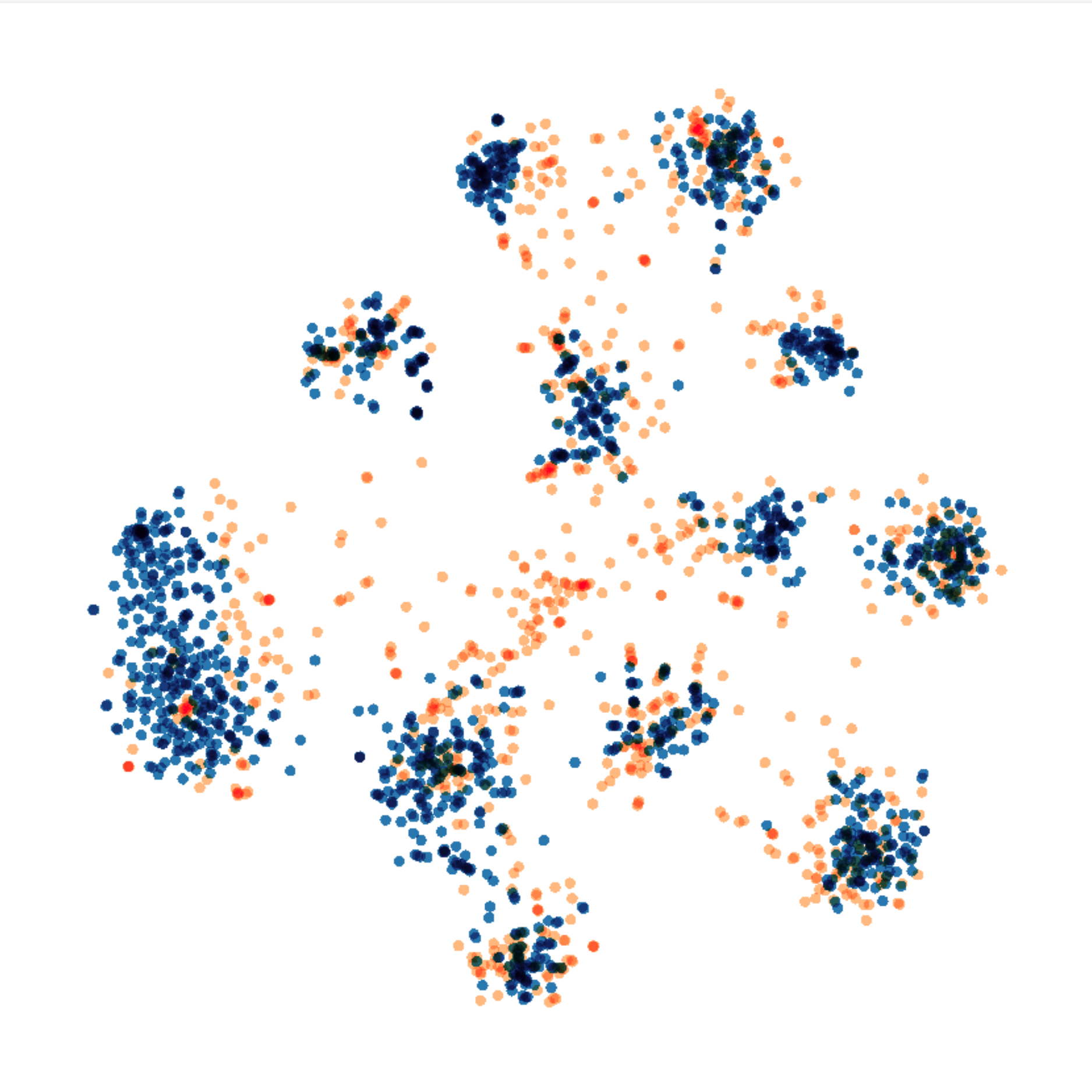}
    \caption{TemPooling + DANN~\cite{ganin2016domain}}
    \label{fig:tSNE_TemPooling_G}
  \end{subfigure}
  \begin{subfigure}[b]{0.235\textwidth}
    \includegraphics[width=\textwidth]{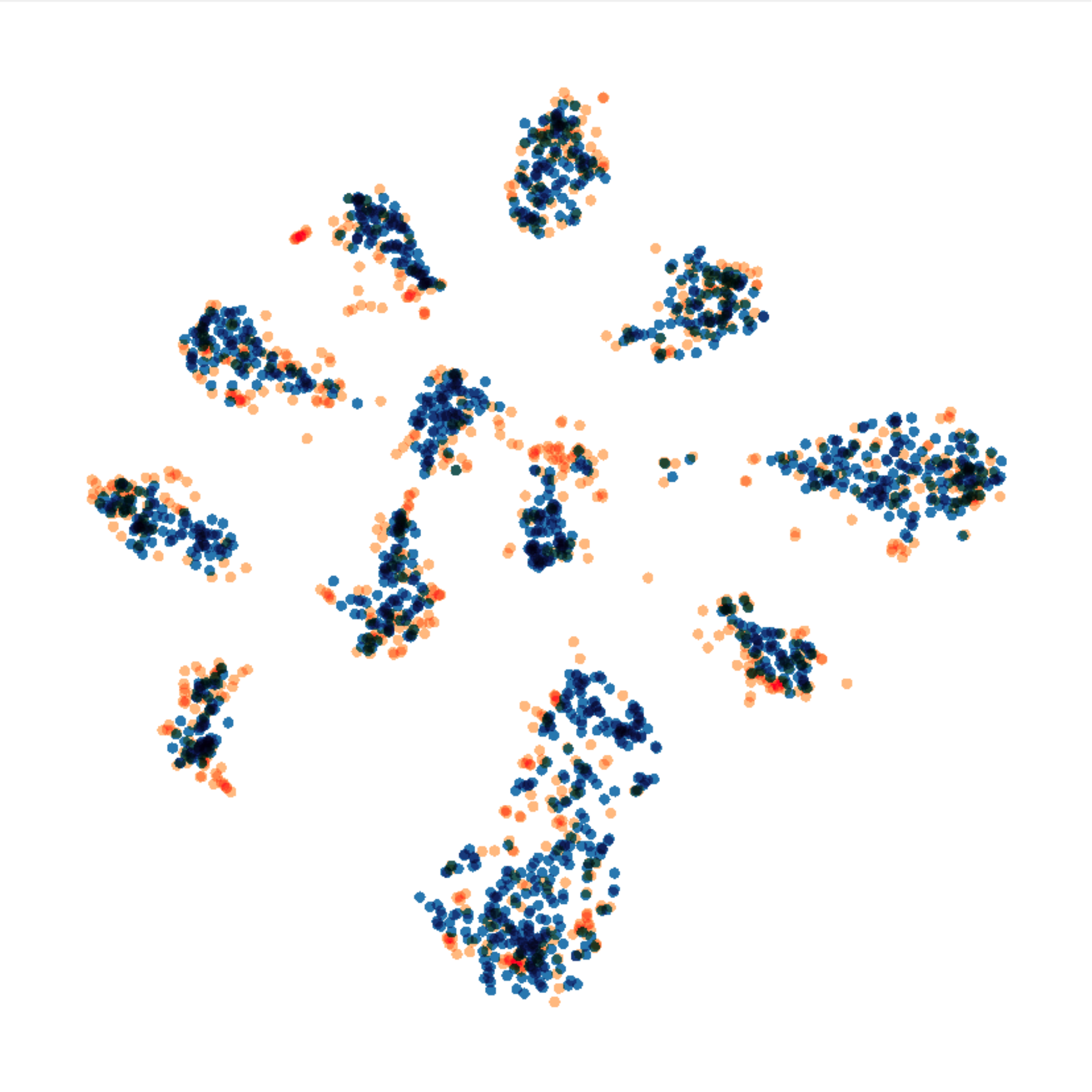}
    \caption{TA$^3$N}
    \label{fig:tSNE_TAAAN}
  \end{subfigure}
\caption{The comparison of t-SNE visualization. The blue dots represent source data while the orange dots represent target data. See the supplementary for more comparison.}
\label{fig:tSNE}
\end{figure}

\textbf{Visualization of distribution}.
To investigate how our approaches bridge the gap between source and target domains, we visualize the distribution of both domains using t-SNE~\cite{maaten2008visualizing}. 
Figure~\ref{fig:tSNE} shows that TA$^3$N can group source data (blue dots) into denser clusters and generalize the distribution into the target domains (orange dots) as well.

\textbf{Domain discrepancy measure}.
To measure the alignment between different domains, we use Maximum Mean Discrepancy (MMD) and domain loss, which are calculated using the final video representations. Lower MMD values and higher domain loss both imply smaller domain gap. TA$^3$N reaches lower discrepancy loss (0.0842) compared to the TemPooling baseline (0.184), and shows great improvement in terms of the domain loss (from 1.116 to 1.9286), as shown in Table~\ref{table:stats}. 

\begin{table}[!t]
\centering
\footnotesize
    \begin{tabular}{c|c|c|c}
     & Discrepancy & Domain & Validation \\ 
     & loss & loss & accuracy \\ \hline
    TemPooling & 0.1840 & 1.1163 & 70.28 \\ 
    TemPooling + DANN~\cite{ganin2016domain} & 0.1604 & 1.2023 & 71.11 \\ 
    TemRelation & 0.2626 & 1.7588 & 71.67 \\ \hline
    TA$^3$N & \textbf{0.0842} & \textbf{1.9286} & \textbf{78.33} \\ \hline
    \end{tabular}
\caption{The discrepancy loss (MMD), domain loss and validation accuracy of our baselines and proposed approaches.}
\label{table:stats}
\end{table}

\section{Conclusion and Future Work}
In this paper, we present two large-scale datasets for video domain adaptation, \textbf{UCF-HMDB$\boldsymbol{_{full}}$} and \textbf{Kinetics-Gameplay}, including both real and virtual domains. 
We use these datasets to investigate the domain shift problem across videos, and show that simultaneously aligning and learning temporal dynamics achieves effective alignment without the need for sophisticated DA methods.
Finally, we propose \textbf{Temporal Attentive Adversarial Adaptation Network (TA$\boldsymbol{^3}$N)} to simultaneously attend, align and learn temporal dynamics across domains, achieving state-of-the-art performance on all of the cross-domain video datasets investigated. The code and data are released \href{http://github.com/cmhungsteve/TA3N}{here}.

The ultimate goal of our research is to solve real-world problems. Therefore, in addition to integrating more DA approaches into our video DA pipelines, there are two main directions we would like to pursue for future work: 1) apply TA$^3$N to different cross-domain video tasks, including video captioning, segmentation, and detection; 2) we would like to extend these methods to the open-set setting~\cite{busto2017open, saito2018open, peng2018syn2real, hsu2018learning}, which has different categories between source and target domains. The open-set setting is much more challenging but closer to real-world scenarios.

\section{Supplementary}
In the supplementary material, we would like to show more detailed ablation studies, more implementation details, and a complete introduction of the datasets.

\subsection{Visualization of distribution}
\begin{figure}[!ht]
\centering
  \begin{subfigure}[b]{0.235\textwidth}
  \centering
    \includegraphics[width=0.855\textwidth]{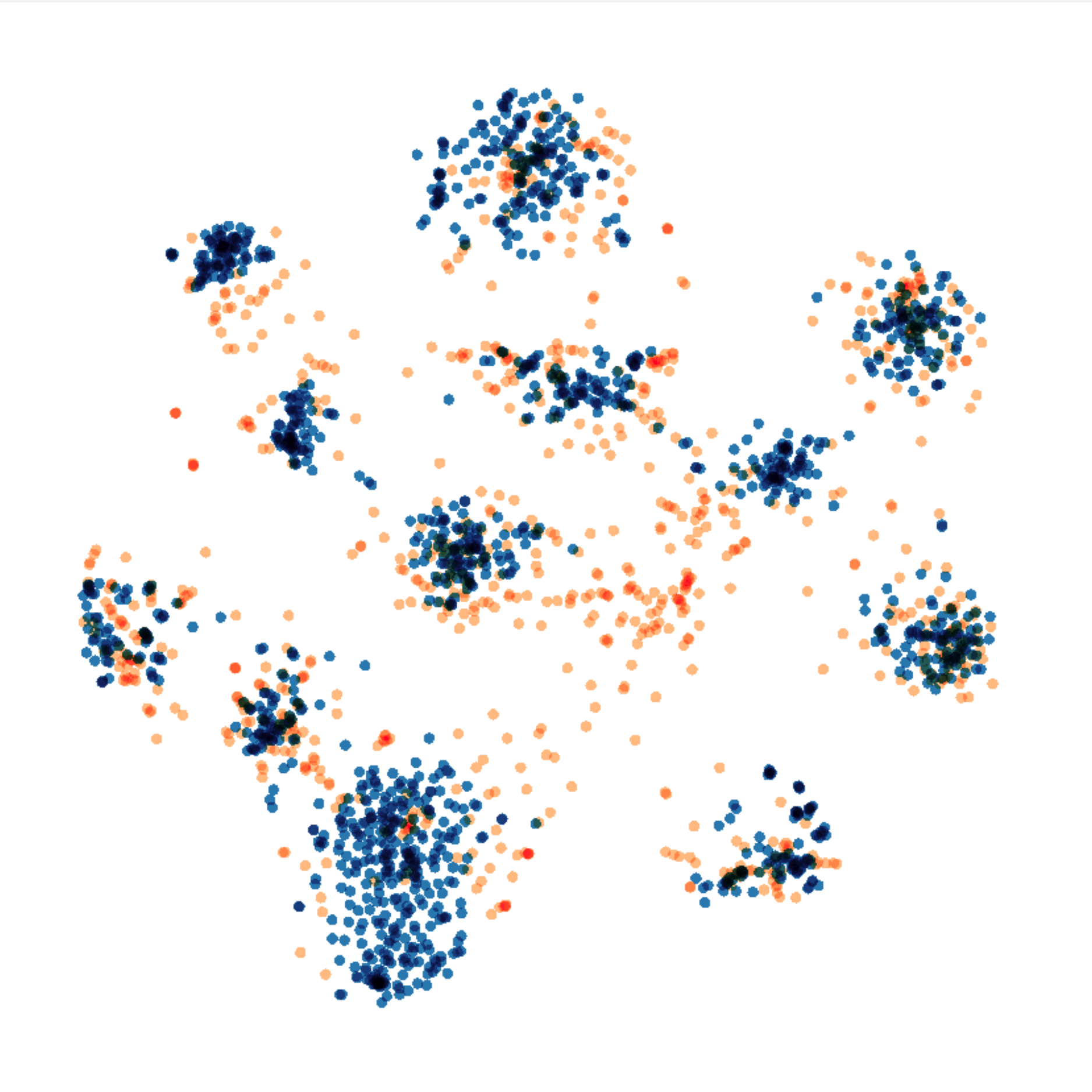}
    \caption{TemPooling}
    \label{fig:tSNE_TemPooling_supp}
  \end{subfigure}
  \begin{subfigure}[b]{0.235\textwidth}
  \centering
    \includegraphics[width=0.855\textwidth]{figs/tSNE/hmdb_ucf-TemPooling-RevGrad_00_05_00.pdf}
    \caption{TemPooling + DANN~\cite{ganin2016domain}}
    \label{fig:tSNE_TemPooling_G_supp}
  \end{subfigure}
  \begin{subfigure}[b]{0.235\textwidth}
  \centering
    \includegraphics[width=0.855\textwidth]{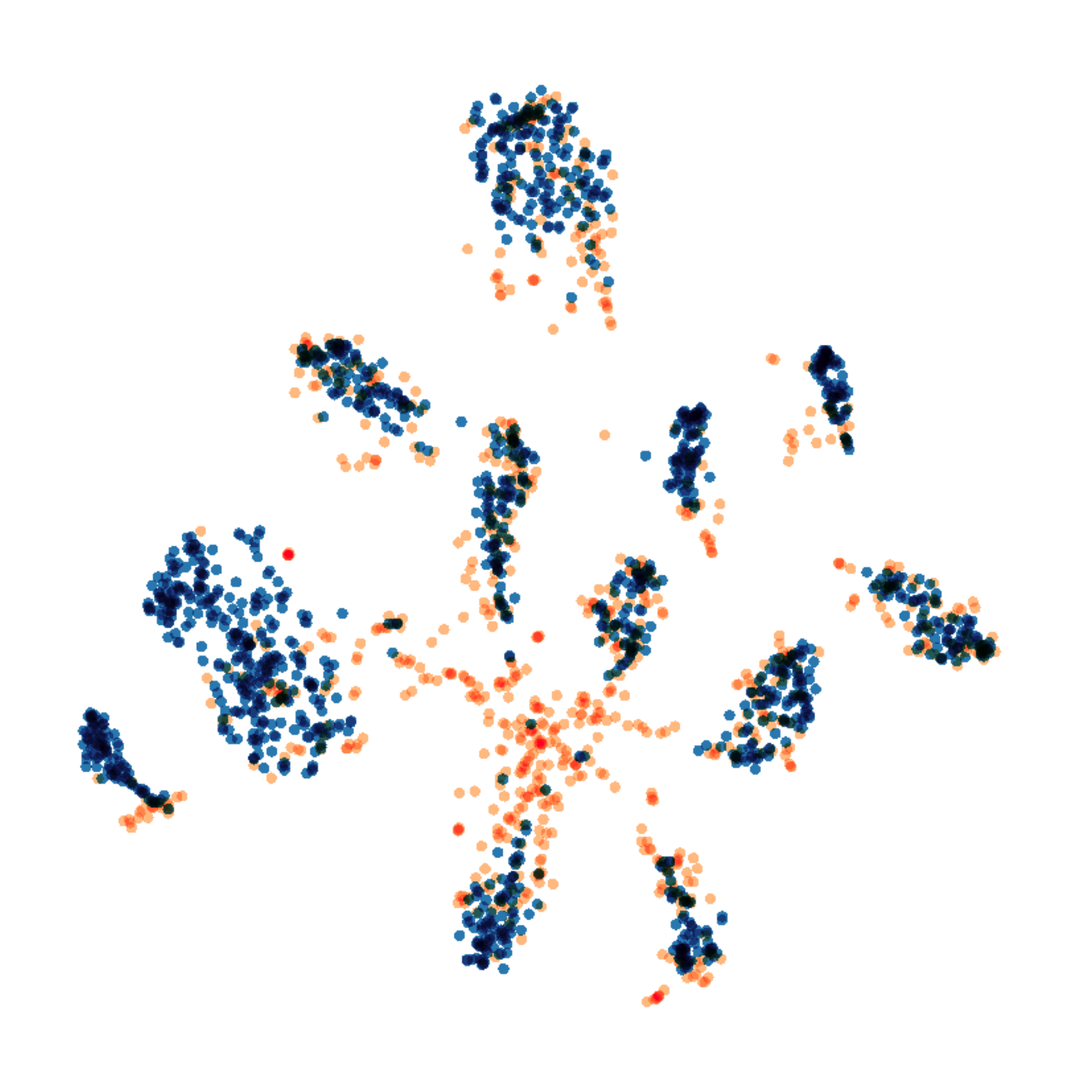}
    \caption{TemRelation}
    \label{fig:tSNE_TemRelation_supp}
  \end{subfigure}
  \begin{subfigure}[b]{0.235\textwidth}
  \centering
    \includegraphics[width=0.855\textwidth]{figs/tSNE/hmdb_ucf-TemRelation-RevGrad_075_075_05-TransAttn-gamma_0003.pdf}
    \caption{TA$^3$N}
    \label{fig:tSNE_TAAAN_supp}
  \end{subfigure}
\caption{The comparison of t-SNE visualization with source (blue) and target (orange) distributions.}
\label{fig:tSNE_supp}
\end{figure}

We visualize the distribution of both domains using t-SNE~\cite{maaten2008visualizing} to investigate how our approaches bridge the gap between the source and target domains.
\Cref{fig:tSNE_TemPooling_supp,fig:tSNE_TemPooling_G_supp} show that the models using the TemPooling architecture poorly align the distribution between different domains, even with the integration of image-based DA approaches. Figure~\ref{fig:tSNE_TemRelation_supp} shows the temporal relation module helps to group source data (blue) into denser clusters but is still not able to generalize the distribution into the target domains (orange). Finally, with TA$^3$N, data from both domains are clustered and aligned with each other (Figure~\ref{fig:tSNE_TAAAN_supp}).

\subsection{Domain Attention Mechanism}
\begin{figure}[!ht]
\centering
\includegraphics[scale=0.41]{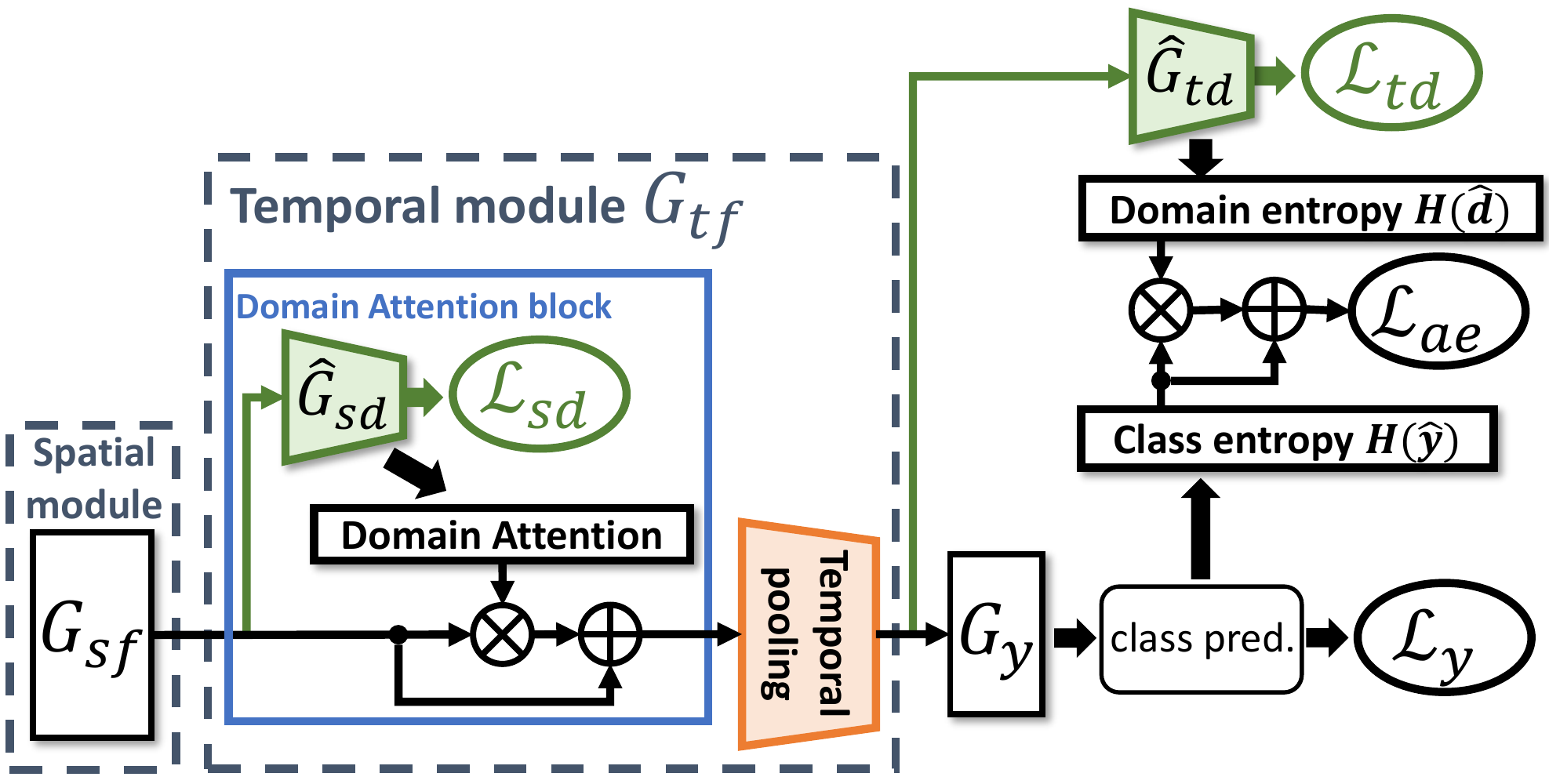}
\caption{Baseline architecture (TemPooling) equipped with the domain attention mechanism (ignoring the input feature parts to save space).}
\label{fig:TemPooling_RevGrad_TransAttn_supp}
\end{figure}

We also apply the domain attention mechanism to TemPooling by attending to the raw frame features, as shown in \Cref{fig:TemPooling_RevGrad_TransAttn_supp}. \Cref{table:experiments_dann_position_ucf_hmdb_supp,table:experiments_dann_position_hmdb_ucf_supp} show that the domain attention mechanism improves the performance for both TemPooling and TemRelation architectures, including all types of adversarial discriminators. 
This implies that video DA can benefit from domain attention even if the backbone architecture does not encode temporal dynamics. 

\begin{table}[!ht]
\centering
\scriptsize
    \begin{tabular}{c|c|c|c|c}
    Temporal & \multirow{2}{*}{TemPooling} & TemPooling & \multirow{2}{*}{TemRelation} & TemRelation \\ 
    Module &  & + Attn. &  & + Attn. \\ \hline
    Target only & \multicolumn{2}{c}{80.56 (-)} & \multicolumn{2}{|c}{82.78 (-)} \\ \hline
    Source only & \multicolumn{2}{c|}{70.28 (-)} & \multicolumn{2}{c}{71.67 (-)} \\ \hline
    $\hat{G}_{sd}$ & 71.11 (0.83) & 71.94 (1.66) & 74.44 (2.77) & 75.00 (3.33) \\ 
    $\hat{G}_{td}$ & 71.11 (0.83) & 72.78 (2.50) & 74.72 (3.05) & 76.94 (5.27) \\ 
    $\hat{G}_{rd}$ & - (-) & - (-) & 76.11 (4.44) & 76.94 (5.27) \\ \hline
    All $\hat{G}_d$ & 71.11 (0.83) & \textbf{73.06} (\textbf{2.78}) & 77.22 (5.55) & \textbf{78.33} (\textbf{6.66}) \\ \hline
    \end{tabular}
\caption{The evaluation of accuracy (\%) for integrating $\hat{G}_d$ in different positions on ``U $\rightarrow$ H" . Gain values are in (). }
\label{table:experiments_dann_position_ucf_hmdb_supp}
\end{table}

\begin{table}[!ht]
\centering
\scriptsize
    \begin{tabular}{c|c|c|c|c}
    Temporal & \multirow{2}{*}{TemPooling} & TemPooling & \multirow{2}{*}{TemRelation} & TemRelation \\ 
    Module &  & + Attn. &  & + Attn.  \\ \hline
    Target only & \multicolumn{2}{|c}{92.12 (-)} & \multicolumn{2}{|c}{94.92 (-)} \\ \hline
    Source only & \multicolumn{2}{c|}{74.96 (-)} & \multicolumn{2}{c}{73.91 (-)} \\ \hline
    $\hat{G}_{sd}$ & 75.13 (0.17) & 77.58 (2.62) & 74.44 (1.05) & 78.63 (4.72) \\ 
    $\hat{G}_{td}$ & 75.13 (0.17) & 78.46 (3.50) & 75.83 (1.93) & 81.44 (7.53)  \\ 
    $\hat{G}_{rd}$ & - (-) & - (-) & 75.13 (1.23) & 78.98 (5.07) \\ \hline
    All $\hat{G}_d$ & 75.13 (0.17) & \textbf{78.46} (\textbf{3.50}) & 80.56 (6.66) & \textbf{81.79} (\textbf{7.88}) \\ \hline
    \end{tabular}
\caption{The evaluation of accuracy (\%) for integrating $\hat{G}_d$ in different positions on ``H $\rightarrow$ U" . Gain values are in (). }
\label{table:experiments_dann_position_hmdb_ucf_supp}
\end{table}

\subsection{Implementation Details}
\subsubsection{Detailed architectures}
\begin{figure}[!ht]
\centering
\includegraphics[scale=0.373]{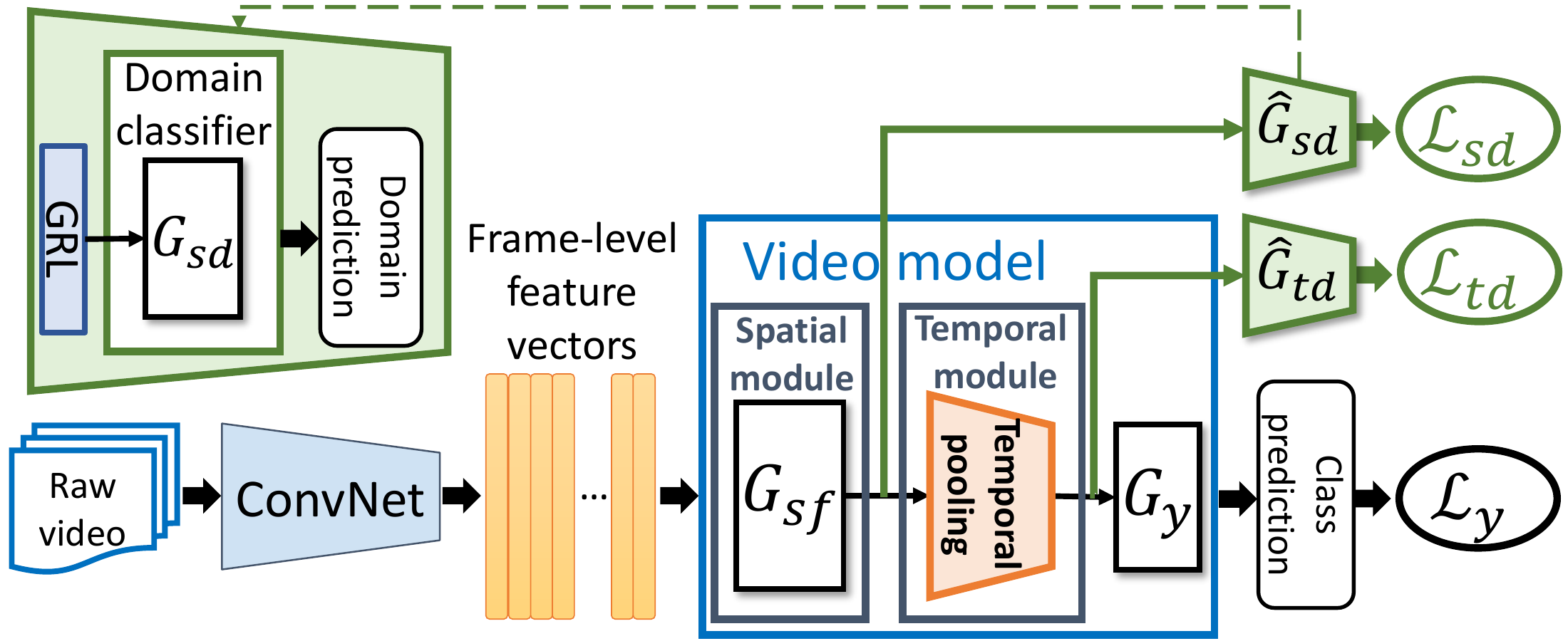}
\caption{The detailed baseline architecture (TemPooling) with the adversarial discriminators $\hat{G}_{sd}$ and $\hat{G}_{td}$.}
\label{fig:TemPooling_RevGrad_supp}
\end{figure}

\begin{figure*}[!t]
\centering
\includegraphics[width=\textwidth]{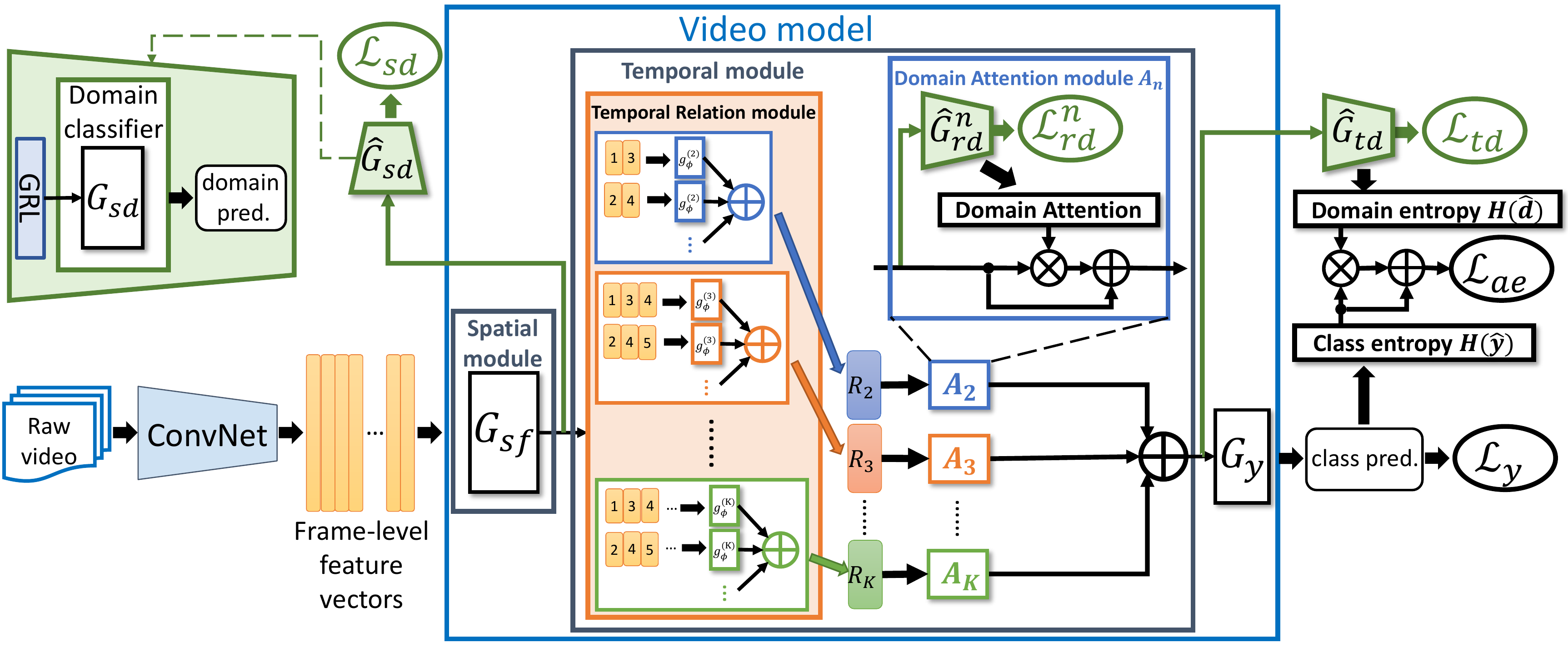}
\caption{The detailed architecture of the proposed TA$^3$N. 
}
\label{fig:TemRelation_RevGrad_TransAttn_supp}
\end{figure*}

The architecture with detailed notations for the baseline is shown in \Cref{fig:TemPooling_RevGrad_supp}. For our proposed TA$^3$N, after generating the $n$-frame relation features $R_n$ by the temporal relation module, we calculate the domain attention value $w^n$ using the domain prediction $\hat{d}$ from the relation discriminator $G^n_{rd}$, and then attend to $R_n$ using $w^n$ with a residual connection. To calculate the attentive entropy loss $\mathcal{L}_{ae}$, since the videos with low domain discrepancy are what we only want to focus on, we attend to the class entropy loss $H(\hat{y})$ using the domain entropy $H(\hat{d})$ as the attention value with a residual connection, as shown in \Cref{fig:TemRelation_RevGrad_TransAttn_supp}. 

\subsubsection{Optimization}
Our implementation is based on the PyTorch~\cite{paszke2017automatic} framework. We utilize the ResNet-101 model pre-trained on ImageNet as the frame-level feature extractor. We sample a fixed number $K$ of frame-level feature vectors with equal spacing in the temporal direction for each video ($K$ is equal to 5 in our setting to limit computational resource requirements).
For optimization, the initial learning rate is $0.03$, and we follow one of the commonly used learning-rate-decreasing strategies shown in DANN~\cite{ganin2016domain}. We use stochastic gradient descent (SGD) as the optimizer with the momentum and weight decay as $0.9$ and $1 \times 10^{-4}$, respectively. 
The ratio between the source and target batch size is proportional to the scale between the source and target datasets. The source batch size depends on the scale of the dataset, which is $32$ for UCF-Olympic and UCF-HMDB$_{small}$, $128$ for UCF-HMDB$_{full}$ and $512$ for Kinetics-Gameplay. 
The optimized values of $\lambda^s$, $\lambda^r$ and $\lambda^t$ are found using the coarse-to-fine grid-search approach. We first search using a coarse-grid with the geometric sequence [0, 10$^{-3}$, 10$^{-2}$, ..., 10$^0$, 10$^1$]. After finding the optimized range of values, [0, 1], we search again using a fine-grid with the arithmetic sequence [0, 0.25, ..., 1]. The final values are 0.75 for $\lambda^s$, 0.5 for $\lambda^r$ and 0.75 for $\lambda^t$, respectively. We search $\gamma$ only by a coarse-grid, and the best value is 0.3.
For future work, we plan to adopt adaptive weighting techniques used for multitask learning, such as uncertainty weighting~\cite{kendall2018multi} and GradNorm~\cite{chen2018gradnorm}, to replace the manual grid-search method.

\subsubsection{Comparison with other work}
As mentioned in the experimental setup, we compare our proposed TA$^3$N with other approaches by extending several state-of-the-art image-based DA methods~~\cite{ganin2016domain, long2017deep, li2018adaptive, saito2018maximum} for video DA with our TemPooling and TemRelation architectures, which are shown as follows:
\begin{enumerate}
    \item \textit{DANN~\cite{ganin2016domain}}: we add one adversarial discriminator $\hat{G}_{sd}$ right after the spatial module and add another one $\hat{G}_{td}$ right after the temporal module. We do not add one more discriminator for relation features for the fair comparison between TemPooling and TemRelation.
    \item \textit{JAN~\cite{long2017deep}}: we add Joint Maximum Mean Discrepancy (JMMD) to the final video representation and the class prediction.
    \item \textit{AdaBN~\cite{li2018adaptive}}: we integrate an adaptive batch-normalization layer into the feature generator $G_{sf}$. In the adaptive batch-normalization layer, the statistics (mean and variance) for both source and target domains are calculated, but only the target statistics are used for validating the target data. 
    \item \textit{MCD~\cite{saito2018maximum}}: we add another classifier $G'_y$ and follow the adversarial training procedure of Maximum Classifier Discrepancy to iteratively optimize the generators ($G_{sf}$ and $G_{tf}$) and the classifier ($G_{y}$).
\end{enumerate}

\subsection{Datasets}
The full summary of all four datasets investigated in this paper is shown in \Cref{table:dataset_supp}.

\begin{table*}[!t]
\centering
\footnotesize
    \begin{tabular}{c|c|c|c|c}
     & UCF-HMDB$_{small}$ & UCF-Olympic & UCF-HMDB$_{full}$ & Kinetics-Gameplay \\ \hline
    length (sec.) & 1 - 21 & 1 - 39 & 1 - 33 & 1 - 10 \\ \hline
    resolution & \multicolumn{4}{c}{UCF: $320\times240$ / Olympic: vary / HMDB: vary$\times240$ / Kinetics: vary / Gameplay: $1280\times720$} \\ \hline
    frame rate & \multicolumn{4}{c}{UCF: 25 / Olympic: 30 / HMDB: 30 / Kinetics: vary / Gameplay: 30} \\ \hline
    class \# & 5 & 6 & 12 & 30 \\ \hline
    training video \# & UCF: 482 / HMDB: 350 & UCF: 601 / Olympic: 250 & UCF: 1438 / HMDB: 840 & Kinetics: 43378 / Gameplay: 2625 \\ \hline
    validation video \# & UCF: 189 / HMDB: 150 & UCF: 240 / Olympic: 54 & UCF: 571 / HMDB: 360 & Kinetics: 3246 / Gameplay: 749 \\ \hline
    \end{tabular}
\caption{The summary of the cross-domain video datasets.}
\label{table:dataset_supp}
\end{table*}

\subsubsection{UCF-HMDB$\boldsymbol{_{full}}$}
\begin{figure}[!t]
\centering
    \begin{subfigure}[b]{0.45\textwidth}
        \includegraphics[width=\textwidth]{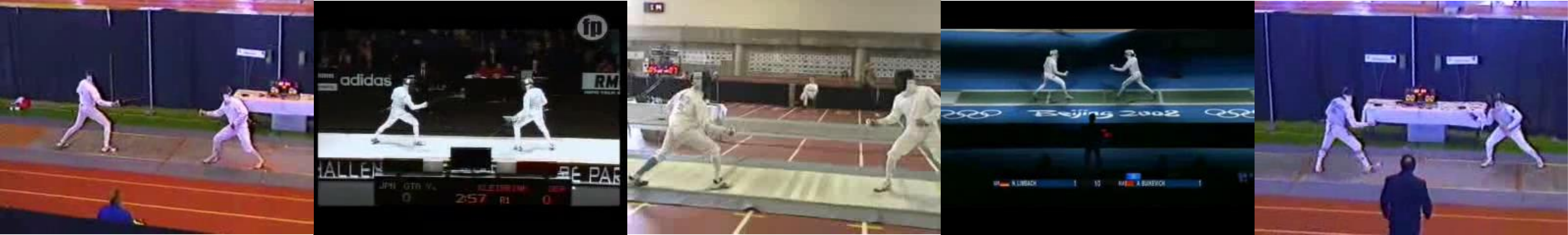}
        \includegraphics[width=\textwidth]{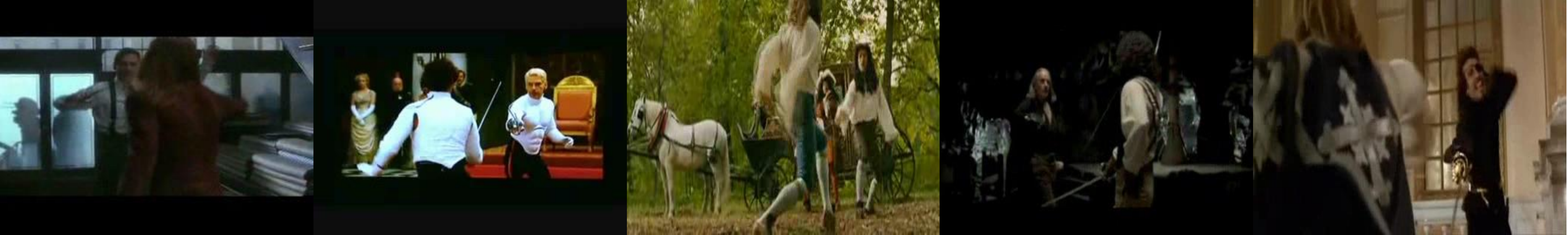}
        \caption{fencing}
        \label{fig:snapshots_ucf_hmdb_fencing_supp}
    \end{subfigure}
    \begin{subfigure}[b]{0.45\textwidth}
        \includegraphics[width=\textwidth]{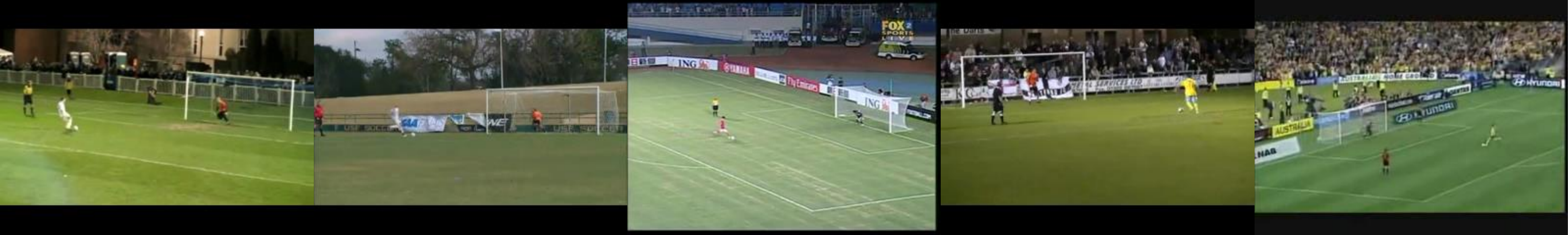}
        \includegraphics[width=\textwidth]{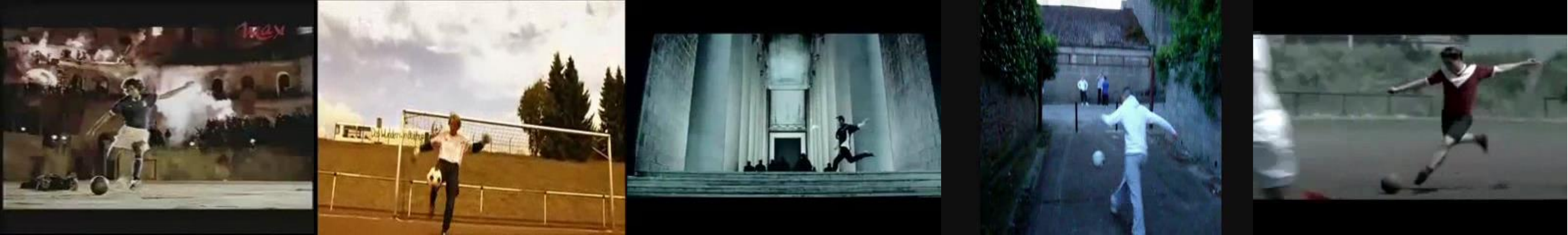}
        \caption{kick\_ball}
        \label{fig:snapshots_ucf_hmdb_kick_ball_supp}
    \end{subfigure}
    \begin{subfigure}[b]{0.45\textwidth}
        \includegraphics[width=\textwidth]{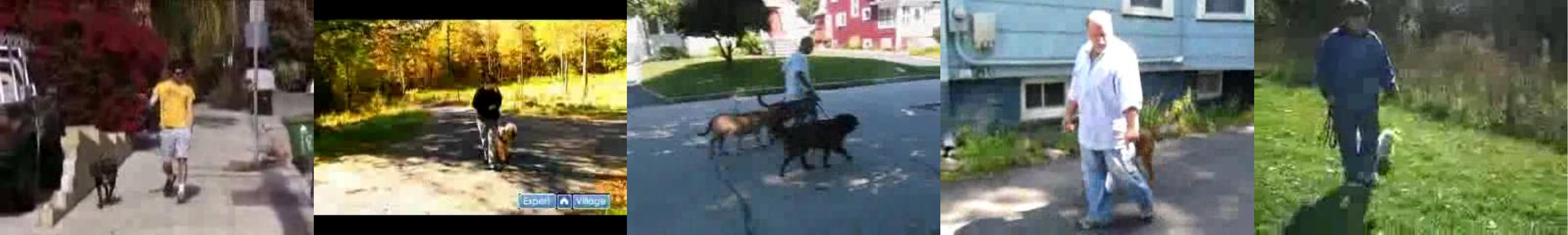}
        \includegraphics[width=\textwidth]{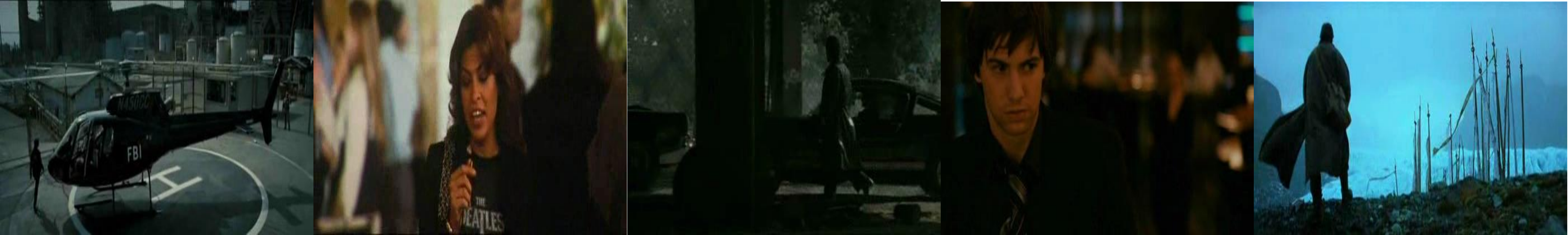}
        \caption{walk}
        \label{fig:snapshots_ucf_hmdb_walk_supp}
    \end{subfigure}
\caption{Snapshots of some example categories on UCF-HMDB$_{full}$. For each category, the snapshots from UCF are shown in the upper row, and the snapshots from HMDB are shown in the lower row.}
\label{fig:snapshots_ucf_hmdb_supp}
\end{figure}

We collect all of the relevant and overlapping categories between UCF101~\cite{soomro2012ucf101} and HMDB51~\cite{kuehne2011hmdb}, which results in 12 categories: \textit{climb, fencing, golf, kick\_ball, pullup, punch, pushup, ride\_bike, ride\_horse, shoot\_ball, shoot\_bow,} and \textit{walk}. Each category may correspond to multiple categories in the original UCF101 or HMDB51 dataset, as shown in \Cref{table:class_ucf_hmdb_supp}. This dataset, \textbf{UCF-HMDB$\boldsymbol{_{full}}$}, includes 1438 training videos and 571 validation videos from UCF, and 840 training videos and 360 validation videos from HMDB, as shown in \Cref{table:dataset_supp}.
Most videos in UCF are from certain scenarios or similar environments,
while videos in HMDB are in unconstrained environments and different camera angles, as shown in \Cref{fig:snapshots_ucf_hmdb_supp}. 

\begin{table}[!t]
\centering
    \begin{tabular}{c|c|c}
    UCF-HMDB$_{full}$ & UCF & HMDB \\ \hline
    climb       & RockClimbingIndoor,   & climb \\
                & RopeClimbing          & \\ \hline
    fencing     & Fencing               & fencing \\ \hline
    golf        & GolfSwing             & golf \\ \hline
    kick\_ball  & SoccerPenalty         & kick\_ball \\ \hline
    pullup      & PullUps               & pullup \\ \hline
    punch       & Punch,                & punch \\ 
                & BoxingPunchingBag,    &  \\
                & BoxingSpeedBag        &  \\ \hline
    pushup      & PushUps               & pushup \\ \hline
    ride\_bike  & Biking                & ride\_bike \\ \hline
    ride\_horse & HorseRiding           & ride\_horse \\ \hline
    shoot\_ball & Basketball            & shoot\_ball \\ \hline
    shoot\_bow  & Archery               & shoot\_bow \\ \hline
    walk        & WalkingWithDog        & walk \\ \hline
    \end{tabular}
\caption{The lists of all collected categories in UCF and HMDB.}
\label{table:class_ucf_hmdb_supp}
\end{table}

\subsubsection{Kinetics-Gameplay}
\begin{figure}[!ht]
\centering
\includegraphics[scale=0.164]{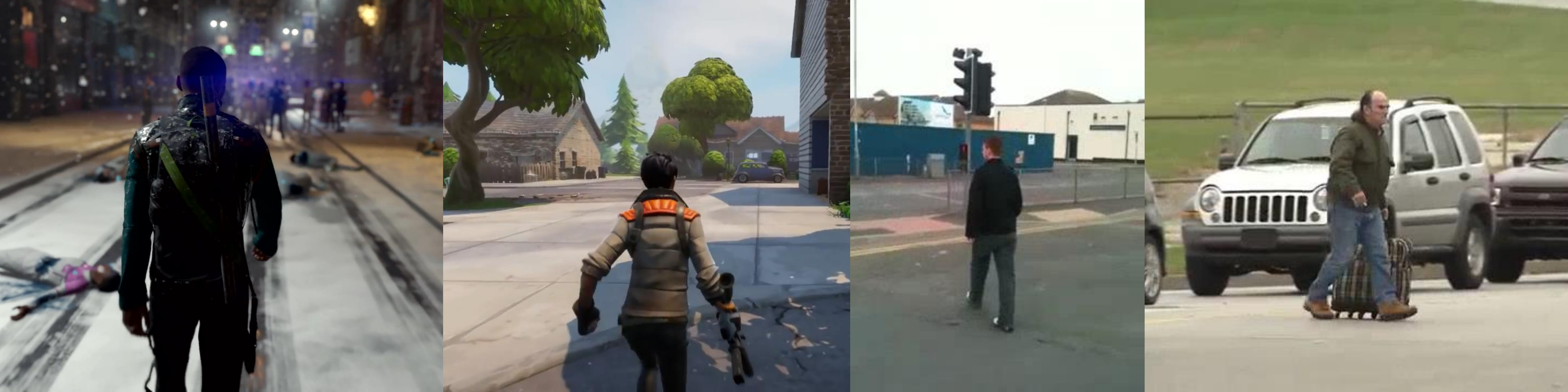}
\caption{Some example screenshots from YouTube videos in Kinetics-Gameplay (left two: Gameplay, right two: Kinetics)}
\label{fig:snapshots_kinetics_gameplay_supp}
\end{figure}

We create the \textit{Gameplay} dataset by first collecting gameplay videos from two video games, \textit{Detroit: Become Human} and \textit{Fortnite}, to build our own action dataset for the virtual domain. The total length of the videos is 5 hours and 41 minutes. We segment all of the raw, untrimmed videos into video clips according to human annotations, which results in 91 categories: \textit{argue, arrange\_object, assemble\_object, break, bump, carry, carve, chop\_wood, clap, climb, close\_door, close\_others, crawl, cross\_arm, crouch, crumple, cry, cut, dance, draw, drink, drive, eat, fall\_down, fight, fix\_hair, fly\_helicopter, get\_off, grab, haircut, hit, hit\_break, hold, hug, juggle\_coin, jump, kick, kiss, kneel, knock, lick, lie\_down, lift, light\_up, listen, make\_bed, mop\_floor, news\_anchor, open\_door, open\_others, paint\_brush, pass\_object, pet, poke, pour, press, pull, punch, push, push\_object, put\_object, raise\_hand, read, row\_boat, run, shake\_hand, shiver, shoot\_gun, sit, sit\_down, slap, sleep, slide, smile, stand, stand\_up, stare, strangle, swim, switch, take\_off, talk, talk\_phone, think, throw, touch, walk, wash\_dishes, water\_plant, wave\_hand,} and \textit{weld}. The maximum length for each video clip is 10 seconds, and the minimum is 1 second. We also split the dataset into training, validation, and testing sets by randomly selecting videos in each category with the ratio 7:2:1. 
We build the \textbf{Kinetics-Gameplay} dataset by selecting 30 overlapping categories between Gameplay and one of the largest public video datasets \textit{Kinetics-600}~\cite{kay2017kinetics, carreira2018kinetics}: \textit{break, carry, clean\_floor, climb, crawl, crouch, cry, dance, drink, drive, fall\_down, fight, hug, jump, kick, light\_up, news\_anchor, open door, paint\_brush, paraglide, pour, push, read, run, shoot\_gun, stare, talk, throw, walk,} and \textit{wash\_dishes}. Each category may also correspond to multiple categories in both datasets, as shown in \Cref{table:class_kinetics_gameplay_supp}. Kinetics-Gameplay includes 43378 training videos and 3246 validation videos from Kinetics, and 2625 training videos and 749 validation videos from Gameplay, as shown in \Cref{table:dataset_supp}. 
Kinetics-Gameplay is much more challenging than UCF-HMDB$_{full}$ due to the significant domain shift between the distributions of virtual and real data. Furthermore, The alignment between imbalanced-scaled source and target data is also another challenge.
Some example snapshots are shown in \Cref{fig:snapshots_kinetics_gameplay_supp}.

\begin{table*}[!t]
\centering
    \begin{tabular}{c|c|c}
    Kinetics-Gameplay & Kinetics & Gameplay \\ \hline
    break           & breaking boards, smashing                                         & break, bump, hit\_break \\ \hline
    carry           & carrying baby                                                     & carry \\ \hline
    clean\_floor    & mopping floor                                                     & mop\_floor \\ \hline
    climb           & climbing a rope, climbing ladder, climbing tree,                   & climb \\ 
                    & ice climbing, rock climbing                                       &  \\ \hline
    crawl           & crawling baby                                                     & crawl \\ \hline
    crouch          & squat, lunge                                                      & crouch, kneel \\ \hline
    cry             & crying                                                            & cry \\ \hline
    dance           & belly dancing, krumping, robot dancing                            & dance \\ \hline
    drink           & drinking shots, tasting beer                                      & drink \\ \hline
    drive           & driving car, driving tractor                                      & drive \\ \hline
    fall\_down      & falling off bike, falling off chair, faceplanting                 & fall\_down \\ \hline
    fight           & pillow fight, capoeira, wrestling,                                & fight, strangle, \\ 
                    & punching bag, punching person (boxing)                            & punch, hit \\ \hline
    hug             & hugging (not baby), hugging baby                                  & hug \\ \hline
    jump            & high jump, jumping into pool,                                     & jump \\ 
                    & parkour                                                           &  \\ \hline
    kick            & drop kicking, side kick                                           & kick \\ \hline
    light\_up       & lighting fire                                                     & light\_fire \\ \hline
    news\_anchor    & news anchoring                                                    & news\_anchor \\ \hline
    open\_door      & opening door, opening refrigerator                                & open\_door \\ \hline
    paint\_brush    & brush painting                                                    & paint\_brush \\ \hline
    paraglide       & paragliding                                                       & paraglide \\ \hline
    pour            & pouring beer                                                      & pour \\ \hline
    push            & pushing car, pushing cart, pushing wheelbarrow,                   & push,  \\ 
                    & pushing wheelchair, push up                                       & push\_object \\ \hline
    read            & reading book, reading newspaper                                   & read \\ \hline
    run             & running on treadmill, jogging                                     & run \\ \hline
    shoot\_gun      & playing laser tag, playing paintball                              & shoot\_gun \\ \hline
    stare           & staring                                                           & stare \\ \hline
    talk            & talking on cell phone, arguing, testifying                        & talk, argue, talk\_phone \\ \hline
    throw           & throwing axe, throwing ball (not baseball or American football),   & throw \\ 
                    & throwing knife, throwing water balloon                            &  \\ \hline
    walk            & walking the dog, walking through snow, jaywalking                 & walk \\ \hline
    wash\_dishes    & washing dishes                                                    & wash\_dishes \\ \hline
    \end{tabular}
\caption{The lists of all collected categories in Kinetics and Gameplay.}
\label{table:class_kinetics_gameplay_supp}
\end{table*}

\subsection{More Details}
\subsubsection{JAN on Kinetics-Gameplay}
JAN~\cite{long2017deep} does not perform well on Kinetics-Gameplay compared to the performance on UCF-HMDB$_{full}$. The main reason is the imbalanced size between the source and target data in Kinetics-Gameplay. The discrepancy loss MMD is calculated using the same number of source and target data (not the case for other types of DA approaches). Therefore, in each iteration, MMD is calculated using parts of the source batch and the whole target batch. This means that the domain discrepancy is reduced only between part of source data and target data during training, so the learned model is still overfitted to the source domain. The discrepancy loss MMD works well when the source and target data are balanced, which is the case for most image DA datasets and UCF-HMDB$_{full}$, but not for Kinetics-Gameplay.

\subsubsection{Comparison with AMLS~\cite{jamal2018deep}}
When evaluating on UCF-HMDB$_{small}$, AMLS~\cite{jamal2018deep} fine-tunes their networks using UCF and HMDB, respectively, before applying their DA approach. Here we only show their results which are fine-tuned with source data, because the target labels should be unseen during training in unsupervised DA settings. For example, we don't compare their results which test on HMDB data using the models fine-tuned with HMDB data since it is not unsupervised DA.

\subsubsection{Other baselines}
3D ConvNets~\cite{tran2015learning} have also been used for extracting video-level feature representations. However, 3D ConvNets consume a great deal of GPU memory, and \cite{tran2018closer} also shows that 3D ConvNets are limited by efficiency and effectiveness issues when extracting temporal information.

Optical-flow extracts the motion characteristics between neighbor frames to compensate for the lack of temporal information in raw RGB frames. In this paper, we focus on attending to the temporal dynamics to effectively align domains even with only RGB frames. We consider optical-flow to be complementary to our method.

\subsubsection{Comparison with literature in other fields}
\noindent\textbf{Cycle-consistency.}
Some papers related to cycle-consistency~\cite{wang2019learning,dwibedi2019temporal} introduce self-supervised methods for learning visual correspondence between images or videos from unlabeled videos. They use cycle-consistency as free supervision to learn video representations. The main difference from our approach is that we explicitly align the feature spaces between source and target domains, while these self-supervised methods aim to learn general representations using only the source domain. We see cycle-consistency as a complementary method that can be integrated into our approach to achieve more effective domain alignment.

\noindent\textbf{Robotics.}
In Robotics, it is a common trend to transfer the models trained in simulation to real world. One of the effective method to bridge the domain gap is randomizing the dynamics of the simulator during training to improve the robustness for different environments~\cite{peng2018sim}.
The setting is different from our task because we focus on feature learning rather than policy learning, and we see domain randomization as a complementary technique that can extend our approach to a more generalized version.

\subsubsection{Failure cases for TemRelation}
TemRelation shows limited improvement over TemPooling for some categories with consistency across time. For example, with the same DA method (DANN), TemRelation has the same accuracy with TemPooling for \textit{ride\_bike} (97\%), and has lower accuracy for \textit{ride\_horse} (93\% and 97\%). The possible reason is that temporal pooling can already model temporally consistent actions well, and it may be redundant to model these actions with multiple timescales like TemRelation.

\subsubsection{Testing time for TA$^3$N}
Different from TA$^2$N, TA$^3$N passes data to all the domain discriminators during testing. However, since all our domain discriminators are shallow, the testing time is similar. In our experiment, TA$^3$N only computes 10\% more time than TA$^2$N.

{\small
\bibliographystyle{ieee_fullname}
\bibliography{egbib}
}

\end{document}